\PassOptionsToPackage{numbers,square,compress}{natbib}
\documentclass[table]{gtech}
\PassOptionsToPackage{table, usenames, dvipsnames}{xcolor}

\RequirePackage{tgpagella}
\RequirePackage{mathpazo}
\RequirePackage{inconsolata}

\usepackage[utf8]{inputenc}
\usepackage{url}
\usepackage{booktabs}
\usepackage{amsfonts}
\usepackage{nicefrac}
\usepackage{microtype}
\usepackage{xcolor}
\usepackage{xspace}

\usepackage{float}
\usepackage{pifont}
\usepackage{bbding}
\usepackage{threeparttable}
\usepackage{colortbl}
\usepackage{graphicx}
\usepackage{makecell}
\usepackage{bbm}
\usepackage{multirow}
\usepackage{amsmath}
\usepackage{amssymb}
\usepackage{adjustbox}
\usepackage{enumitem}
\usepackage{textcomp}
\usepackage{hypernat}
\usepackage{bookmark}

\newcommand{\vpara}[1]{\vspace{0.05in}\noindent\textbf{#1 }}
\newcommand{\corrdag}{\textsuperscript{\textdagger}}

\definecolor{njupurple}{HTML}{6A005F}
\hypersetup{
    colorlinks=true,
    linkcolor=njupurple,
    citecolor=njupurple,
    urlcolor=njupurple,
    citebordercolor=njupurple,
    pdfborder={0 0 0}
}
\AtBeginDocument{%
  \hypersetup{linkcolor=njupurple,citecolor=njupurple,urlcolor=njupurple}%
  \setcitestyle{numbers,square,comma}%
}

\definecolor{colorBest}{RGB}{255, 204, 153}
\definecolor{colorSecond}{RGB}{255, 235, 204}
\newcommand{\res}[2]{#1{\scriptsize$\pm$#2}}
\newcommand{\best}[2]{\cellcolor{colorBest}\textbf{#1}{\scriptsize$\pm$#2}}
\newcommand{\second}[2]{\cellcolor{colorSecond}#1{\scriptsize$\pm$#2}}

\definecolor{colorBestEnc}{RGB}{255, 204, 153}
\definecolor{colorSecondEnc}{RGB}{255, 235, 204}
\newcommand{\bestenc}[1]{\cellcolor{colorBestEnc}\textbf{#1}}
\newcommand{\secondenc}[1]{\cellcolor{colorSecondEnc}{#1}}

\definecolor{colorBestTab}{RGB}{255, 204, 153}
\newcommand{\besttab}[1]{\cellcolor{colorBestTab}\textbf{#1}}

\newenvironment{ack}{%
  \par\smallskip\noindent\textbf{Acknowledgments.}\ignorespaces
}{%
  \par\smallskip
}

\renewcommand\authorformat[2][]{{\fontfamily{optimistic}\selectfont \bfseries #2}}
\renewcommand\author[1]{\addtolist[]{#1}{\authorlist}{\authorformat}{, }}
\renewcommand\contributionformat[2][]{{\small #2}}
\renewcommand\contribution[1]{\addtolist[]{#1}{\contributionlist}{\contributionformat}{, }}

\renewcommand{\title}[1]{\newcommand{\titlelist}{{\huge\fontfamily{optimistic}\selectfont #1}}}

\title{%
\Large{\textcolor{njupurple}{\textbf{\texttt{PyAGC}}}} \\[0.35em]
\huge{Bridging Academia and Industry: A Comprehensive Benchmark for Attributed Graph Clustering}%
}

\author{Yunhui Liu\textsuperscript{1,2}}
\author{Pengyu Qiu\textsuperscript{2}}
\author{Yu Xing\textsuperscript{1}}
\author{Peng Du\textsuperscript{2}}
\author{Yongchao Liu\textsuperscript{2,\textdagger}}
\author{Chuntao Hong\textsuperscript{2}}
\author{Jiajun Zheng\textsuperscript{2}}
\author{Tao Zheng\textsuperscript{1}}
\author{Tieke He\textsuperscript{1,\textdagger}}

\affiliation[1]{State Key Laboratory for Novel Software Technology, Nanjing University}
\affiliation[2]{Ant Group}

\contribution{\corrdag{}Corresponding authors.}

\abstract{%
Attributed Graph Clustering (AGC) is a fundamental unsupervised task that integrates structural topology and node attributes to uncover latent patterns in graph-structured data.
Despite its significance in industrial applications such as fraud detection and user segmentation, a significant chasm persists between academic research and real-world deployment.
Current evaluation protocols suffer from the small-scale, high-homophily citation datasets, non-scalable full-batch training paradigms, and a reliance on supervised metrics that fail to reflect performance in label-scarce environments.
To bridge these gaps, we present \texttt{PyAGC}, a comprehensive, production-ready benchmark and library designed to stress-test AGC methods across diverse scales and structural properties.
We unify existing methodologies into a modular Encode-Cluster-Optimize framework and, for the first time, provide memory-efficient, mini-batch implementations for a wide array of state-of-the-art AGC algorithms.
Our benchmark curates 12 diverse datasets, ranging from $2.7 \times 10^3$ to $1.1 \times 10^8$ nodes, specifically incorporating industrial graphs with complex tabular features and low homophily.
Furthermore, we advocate for a holistic evaluation protocol that mandates unsupervised structural metrics and efficiency profiling alongside traditional supervised metrics.
Battle-tested in high-stakes industrial workflows at Ant Group, this benchmark offers the community a robust, reproducible, and scalable platform to advance AGC research towards realistic deployment.
The code and resources are publicly available via \href{https://github.com/Cloudy1225/PyAGC}{\texttt{GitHub}}, \href{https://pypi.org/project/pyagc}{\texttt{PyPI}}, and \href{https://pyagc.readthedocs.io}{\texttt{Docs}}.
We invite readers to our \href{https://arxiv.org/abs/2603.20829}{latest survey} for the unified framework and industrial perspective that complement this benchmark.%
}

\date{Sep 26, 2026\vspace{-1mm}}
\gtechdata[Venue]{NeurIPS 2026 (Evaluations \& Datasets Track)\vspace{-1mm}}
\gtechdata[Code]{\url{https://github.com/Cloudy1225/PyAGC}\vspace{-1mm}}
\gtechdata[Package]{\url{https://pypi.org/project/pyagc}\vspace{-1mm}}
\gtechdata[Documentation]{\url{https://pyagc.readthedocs.io}\vspace{-1mm}}
\correspondence{%
\email{lyhcloudy1225@gmail.com},
\email{yongchao.ly@antgroup.com},
\email{hetieke@gmail.com}%
}

\begin{document}
\maketitle

\section{Introduction}
\vspace{0.35em}
\noindent\textit{Graph learning will lose relevance due to poor benchmarks.}\hfill
{\small --- Bechler-Speicher et al.~\cite{Position}}\par
\vspace{0.55em}
Attributed Graph Clustering (AGC), which partitions nodes by synergizing structural topology and node attributes, is a pivotal unsupervised learning task across diverse domains~\cite{DGCSurvey, AGCSurvey}. 
Unlike traditional community detection~\cite{Modularity} or KMeans~\cite{KMeans}, AGC must integrate heterogeneous signals to uncover latent patterns, making it particularly valuable in label-scarce industrial applications such as fraud detection~\cite{GFDN}, social analysis~\cite{CGC}, and user segmentation~\cite{ELCRec}.

Recent advances in Graph Neural Networks (GNNs)~\cite{GCN} and Self-Supervised Learning (SSL)~\cite{GSSLSurvey} have rapidly expanded the AGC landscape. 
Existing methods can be unified under an \emph{Encode--Cluster--Optimize} paradigm, which dissects existing algorithms into representation learning, cluster assignment, and optimization objectives. 
Encoders range from non-parametric spectral filters~\cite{AGC, SASE} to parametric GNNs~\cite{DCRN, S3GC}, while clustering strategies include prototype-based models~\cite{DAEGC, DinkNet}, subspace clustering~\cite{SAGSC, MS2CAG}, and differentiable pooling approaches~\cite{DMoN, Neuromap}. 
Optimization schemes further evolve from two-stage pipelines to end-to-end training with graph reconstruction~\cite{GAE, DAEGC}, contrastive learning~\cite{NS4GC, MAGI}, feature decorrelation~\cite{CCASSG, SSGE}, or clustering-specific objectives~\cite{DEC, DinkNet}.

Despite these algorithmic strides, a chasmic gap remains between academic research and industrial reality. 
Through a rigorous survey of existing literature, we identify four critical limitations of the current evaluation ecosystem for AGC as follows:

\begin{itemize}[leftmargin=*]
\item \textbf{The Cora-fication of Datasets:} Most AGC models are evaluated on small citation networks (e.g., \texttt{Cora}, \texttt{CiteSeer}, \texttt{PubMed})~\cite{Position} with clean textual features and high homophily. This creates a path dependency where algorithms are over-fitted to clean academic scenarios, failing to generalize to industrial graphs (e.g., transaction or web networks) that are large-scale, heterophilous, noisy, and dominated by heterogeneous tabular features~\cite{GraphLand}.

\item \textbf{The Scalability Bottleneck:} Many models rely on full-batch operations (e.g., constructing adjacency matrices~\cite{DAEGC, RGAE} or pairwise contrastive learning~\cite{DCRN, NS4GC}) that scale quadratically with node count, restricting applicability to graphs with at most a few hundred thousand nodes. Production-ready mini-batch compatible implementations remain largely absent, leaving the behavior of deep clustering on industrial-scale graphs largely unexplored.

\item \textbf{The Supervised Metric Paradox:} Although AGC is fundamentally an unsupervised task, the community defaults to supervised evaluation metrics such as Accuracy, NMI, and ARI~\cite{DGCSurvey, DGCBench}, creating a label-fitting incentive where models are tuned to recover human-annotated classes rather than discovering intrinsic clusters. In realistic industrial settings where labels are unavailable, practitioners may require algorithms optimizing intrinsic structural density~\cite{EvalComm}, yet these unsupervised quality metrics are systematically underreported.

\item \textbf{The Reproducibility and Benchmarking Gap:} While DGCBench~\cite{DGCBench} has attempted to standardize AGC evaluation via a unified codebase, it largely inherits the traditional academic paradigm, retaining full-batch training and focusing on smaller-scale datasets. Consequently, the critical last-mile challenges of industrial application, specifically massive scalability via mini-batch training, robustness to feature/structure heterogeneity, and purely unsupervised structural validation, remain open problems that require a dedicated benchmark.
\end{itemize}

\begin{figure*}[t]
    \centering
    \includegraphics[width=\linewidth]{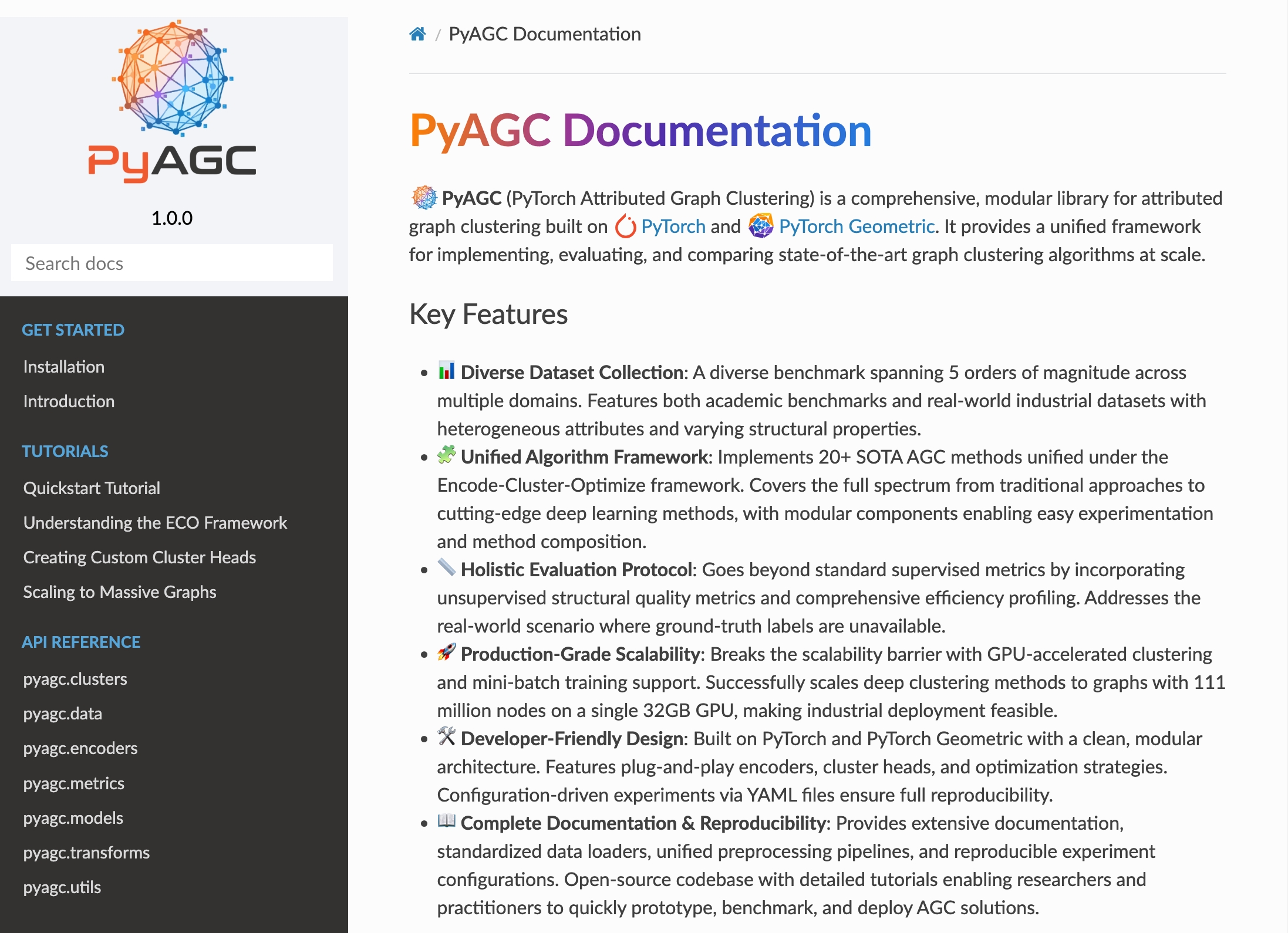}
    \caption{\href{https://pyagc.readthedocs.io}{\texttt{PyAGC} documentation homepage.}}
    \label{fig:pyagc_docs}
\end{figure*}

To bridge these gaps, we introduce a comprehensive benchmark and evaluation framework designed to rigorously stress-test AGC methods across diverse domains, scales, and structural properties. 
Our work is anchored by \texttt{PyAGC}, an industrial-scale library that unifies the clustering landscape under the Encode-Cluster-Optimize framework.
As shown in Figure~\ref{fig:pyagc_docs}, \texttt{PyAGC} is released as a public and user-friendly software stack with unified documentation, codebase, and package support, lowering the barrier for reproducible evaluation and practical adoption.

By addressing the above challenges, our benchmark serves as a bridge between academic innovation and industrial application, setting a new standard for the evaluation of Attributed Graph Clustering.
Crucially, \texttt{PyAGC} has been battle-tested in high-stakes industrial environments at \textbf{Ant Group}, supporting critical workflows in Fraud Detection, Anti-Money Laundering, and User Profiling. 
Our contributions are summarized as follows:

\begin{itemize}[leftmargin=*]
\item \textbf{A Diverse Data Atlas:} We curate 12 datasets spanning $2.7{\times}10^3$ to $1.1{\times}10^8$ nodes across citation, social, e-commerce, and web domains. We deliberately introduce industrial datasets (\texttt{HM}, \texttt{Pokec}, \texttt{WebTopic})~\cite{GraphLand} that stress-test models with low homophily, tabular attributes, and massive scale.

\item \textbf{Production-Grade Scalability:} We provide standardized mini-batch implementations for diverse AGC algorithms, demonstrating that methods previously limited to small graphs can scale to 111 million nodes with promising performance, paving the way for industrial adoption.

\item \textbf{Holistic Evaluation Protocol:} Beyond supervised metrics, we mandate unsupervised structural metrics (Modularity~\cite{Modularity}, Conductance~\cite{Conductance}) to assess intrinsic cluster quality in the absence of labels and comprehensive efficiency profiling covering training time and memory consumption.

\item \textbf{A Unified, Reproducible Framework:} We release \texttt{PyAGC}, a modular PyTorch library that unifies data processing, model initialization, and evaluation, enabling the training of deep clustering models on graphs with over 111 million nodes in under 2 hours on a single 32GB V100 GPU.
\end{itemize}

\section{Preliminaries}
\label{sec:preliminaries}

\subsection{Problem Formulation}
Let $\mathcal{G} = (\mathcal{V}, \mathcal{E}, \boldsymbol{X})$ denote a graph with $N$ nodes, adjacency matrix $\boldsymbol{A} \in \{0,1\}^{N \times N}$, and node features $\boldsymbol{X} \in \mathbb{R}^{N \times D}$. 
The goal of attributed graph clustering is to partition nodes into $K$ clusters $\mathcal{C} = \{C_1,\dots,C_K\}$ by jointly leveraging topology and attributes without labels. 
This is typically formulated as learning a mapping $\mathcal{F}: (\boldsymbol{A}, \boldsymbol{X}) \rightarrow \boldsymbol{P} \in [0,1]^{N \times K}$, where $\boldsymbol{P}_{ik}$ denotes the assignment probability of node $v_i$ to cluster $C_k$. 
\emph{Graph Homophily}~\cite{ACMGNN} can be measured by edge homophily $\mathcal{H}_{e}$ (the proportion of edges that connect two nodes in the same class) and node homophily $\mathcal{H}_{n}$ (the average proportion of edge-label consistency of all nodes).

\vspace{-0.2cm}
\subsection{The Encode-Cluster-Optimize Framework}
To organize the disparate landscape of AGC algorithms, we unify them under the \textbf{Encode-Cluster-Optimize (ECO)} framework, which decomposes the clustering process into three distinct modules: Representation Encoding, Cluster Projection, and Optimization Strategy. 
Table~\ref{tab:eco_taxonomy} provides a detailed classification of representative methods within this framework.

\vpara{The Representation Encoding Module ($\mathcal{E}$).}
The encoder $\mathcal{E}$ is responsible for fusing structural topology and node attributes into a latent representation space $\boldsymbol{Z} \in \mathbb{R}^{N \times H}$:
\begin{equation}
    \boldsymbol{Z} = \mathcal{E}(\boldsymbol{A}, \boldsymbol{X}; \Theta_{\mathcal{E}}),
\end{equation}
where $\Theta_{\mathcal{E}}$ denotes the learnable parameters. 
Existing methods diverge in their parameterization of $\mathcal{E}$:
\begin{itemize}[leftmargin=*]
    \item \textit{Parametric Encoders:} Typically instantiated as graph neural networks such as GCN~\cite{GCN} or GAT~\cite{GAT}. These models learn complex non-linear mappings by fusing topology and attributes via message passing~\cite{DAEGC, S3GC, MAGI}.

    \item \textit{Non-Parametric Encoders:} These involve fixed filtering operations that smooth features over the graph topology without training weights, such as low-pass graph filters or adaptive smoothing~\cite{AGC, NAFS, MS2CAG}. In this case, $\Theta_{\mathcal{E}} = \emptyset$.
\end{itemize}

\vpara{The Cluster Projection Module ($\mathcal{C}$).}
The clusterer $\mathcal{C}$ transforms the latent embeddings $\boldsymbol{Z}$ into a cluster assignment matrix $\boldsymbol{P} \in \mathbb{R}^{N \times K}$, representing either soft probabilities or hard partitions:
\begin{equation}
    \boldsymbol{P} = \mathcal{C}(\boldsymbol{Z}; \Theta_{\mathcal{C}}).
\end{equation}
Based on the differentiability of this mapping, which dictates the feasibility of end-to-end optimization, we categorize clustering mechanisms into two fundamental paradigms:

\begin{itemize}[leftmargin=*]
    \item \textit{Differentiable Projection:} 
    This category treats clustering as a differentiable layer within a neural network, allowing gradients to flow from the loss back to the encoder. 
    It encompasses discriminative approaches~\cite{DMoN, Neuromap} that use Softmax-activated MLPs/GNNs to output probabilities directly, and prototypical approaches~\cite{DAEGC, DinkNet} that compute soft assignments via differentiable kernels (e.g., Student's $t$-distribution~\cite{DEC}) over learnable prototypes.
    
    \item \textit{Discrete / Post-hoc Projection:} 
    This category applies a discrete clustering algorithm to the fixed representations generated by the encoder. 
    Since the mapping is non-differentiable, the clustering process is decoupled from representation learning. 
    Common instantiations include KMeans~\cite{NS4GC, MAGI}, Spectral Clustering~\cite{AGC, SASE}, and Subspace Clustering~\cite{SAGSC, MS2CAG}. 
\end{itemize}

\vpara{The Optimization Strategy ($\mathcal{O}$).}
The optimization strategy defines the objective function and the interaction between encoding $\Theta_{\mathcal{E}}$ and clustering $\Theta_{\mathcal{C}}$. 
The general objective is formulated as:
\begin{equation}
    \min_{\Theta_{\mathcal{E}}, \Theta_{\mathcal{C}}} \mathcal{L}_{\text{total}} = \mathcal{L}_{\text{rep}}(\boldsymbol{A}, \boldsymbol{X}, \boldsymbol{Z}) + \lambda \mathcal{L}_{\text{clust}}(\boldsymbol{P}, \boldsymbol{A}).
\end{equation}
where $\mathcal{L}_{\text{rep}}$ supervises the quality of embeddings (e.g., via graph reconstruction~\cite{GAE, GraphMAE} or contrastive loss~\cite{NS4GC, MAGI}) and $\mathcal{L}_{\text{clust}}$ enforces clustering structure (e.g., KL divergence~\cite{DEC}, Modularity maximization~\cite{DMoN}, or Dilation+Shrink~\cite{DinkNet}).

Crucially, the coordination of these components defines two distinct training paradigms:
\begin{itemize}[leftmargin=*]
    \item \textit{Decoupled Training:} The encoder is pre-trained using only the self-supervised objective $\mathcal{L}_{\text{rep}}$, followed by a disjoint application of the post-hoc clusterer~\cite{BLNN, NS4GC, MAGI}. 
    \item \textit{Joint Training:} The encoder and clusterer are optimized end-to-end~\cite{DAEGC, DinkNet, DMoN}. The clustering objective $\mathcal{L}_{\text{clust}}$ provides supervisory signals to refine the embeddings.
\end{itemize}


\begin{figure*}[t]
\centerline{\includegraphics[width=1.\linewidth]{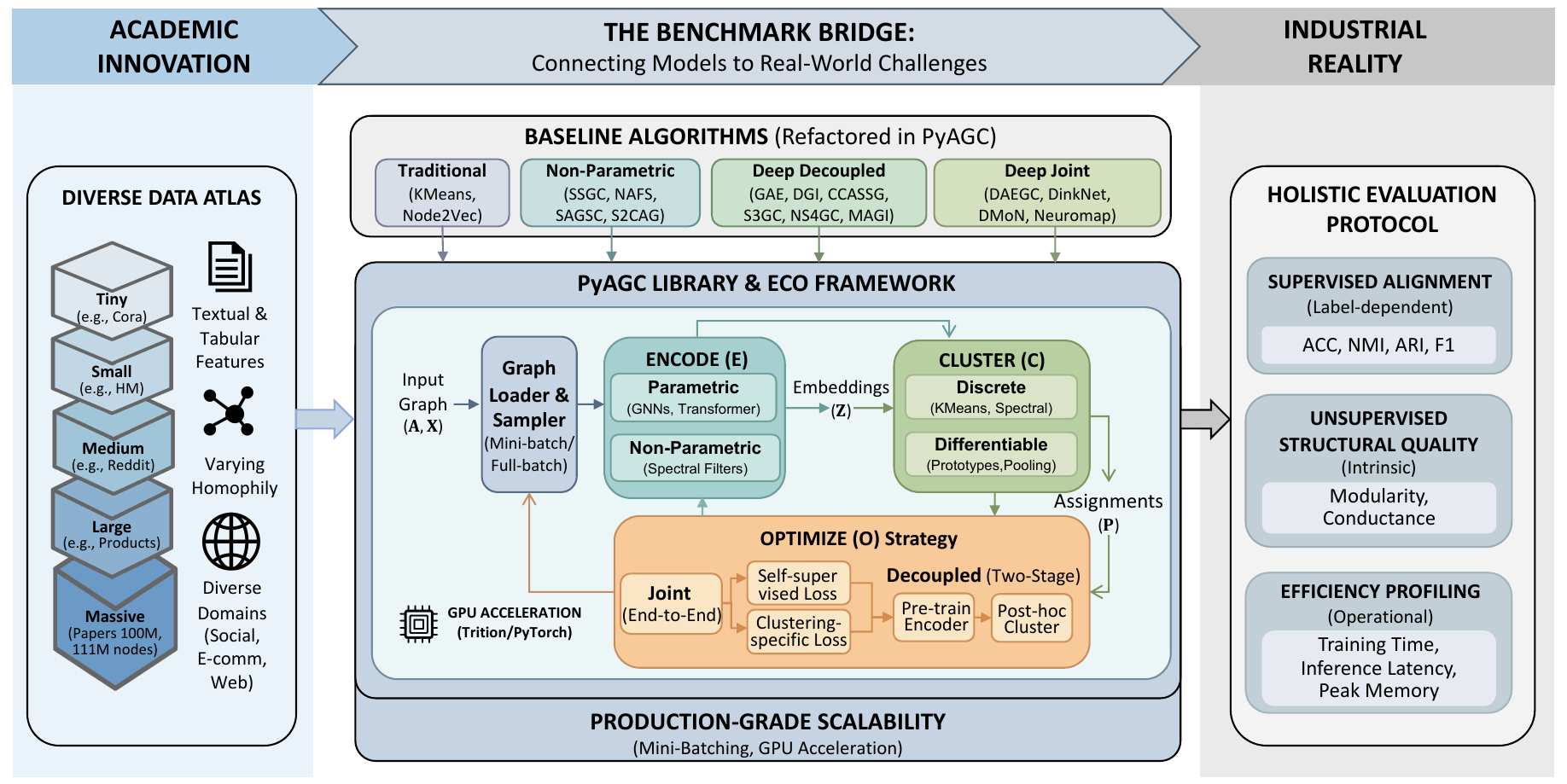}}
\caption{Overview of our benchmark. 
It unifies (1) a diverse data atlas spanning academic and industrial domains, (2) the modular Encode-Cluster-Optimize methodology with memory-efficient mini-batch support, and (3) a holistic evaluation protocol covering supervised alignment, structural quality, and efficiency.
}\label{fig:framework}
\end{figure*}

\section{The PyAGC Library}
\label{sec:pyagc}
To operationalize the ECO taxonomy and resolve the reproducibility crisis in graph clustering, we introduce \texttt{PyAGC}, a modular, production-grade library designed for extensibility and massive scalability. 
The library structure is visualized in Figure~\ref{fig:framework}.

\subsection{Modular Architecture}
\texttt{PyAGC} decouples the clustering lifecycle into interchangeable modules. This design allows researchers to rapidly prototype new methods by swapping components without rewriting the training loop:

\begin{itemize}[leftmargin=*]
    \item \textbf{Encoders (\texttt{pyagc.encoders}):} 
    This module houses the representation learning backbones. 
    We follow \cite{TunedGNN} to provide optimized implementations of standard GNNs~\cite{GCN, SAGE, GAT, GIN} that support both full-batch processing for small graphs and neighbor-sampling-based mini-batching for massive graphs. 
    We also support graph transformers~\cite{SGFormer, Polynormer} implemented by PyTorch Geometric~\cite{PyG}. 
    This allows any clustering head to be easily paired with varying backbones without code duplication.
    
    \item \textbf{Clusters (\texttt{pyagc.clusters}):} 
    We refactor clustering mechanisms into standalone \textit{Cluster Heads}. 
    Whether it is the differentiable pooling~\cite{DMoN, Neuromap}, the prototypical assignment~\cite{DAEGC, DinkNet}, or the discrete partitioning~\cite{KMeans}, each module inherits from a base class ensuring consistent input/output interfaces. 
    This abstraction allows a single encoder to be easily paired with different clustering mechanisms to isolate performance gains.
    
    \item \textbf{Models (\texttt{pyagc.models}):} The high-level model classes orchestrate the interaction between the encoder and the cluster head. 
    They define the optimization strategy, managing the forward pass and the computation of joint losses (e.g., reconstruction loss + clustering loss).
    This design facilitates plug-and-play experimentation; for instance, swapping a GCN encoder for a GAT within the DAEGC framework requires changing a single line in the configuration file.

    \item \textbf{Unified Evaluation:} 
    \texttt{PyAGC} integrates data processing, metric computation, and experiment management. 
    It provides unified data loaders and graph augmentations, alongside an optimized suite of label-based and structural metrics. 
    All experiments are strictly configuration-driven via YAML files, decoupling hyperparameters from implementation to ensure full reproducibility and a seamless transition between different graph datasets.
\end{itemize}

\subsection{Engineering for Efficiency}
To mitigate the computational barriers of deep graph clustering, \texttt{PyAGC} incorporates two critical design optimizations targeting speed and memory efficiency:

\textbf{1. GPU-Accelerated Clustering:} 
Standard CPU-based implementations (e.g., \texttt{Sklearn}) of KMeans are strictly single-threaded and become prohibitively slow when $N > 10^6$.
To resolve this, we implement GPU-accelerated clustering modules using PyTorch and OpenAI Triton\footnote{\url{https://github.com/triton-lang/triton}}, which achieve multi-fold speedups over CPU counterparts, enabling faster clustering during training or testing.

\textbf{2. Mini-Batch Training Protocol:}
We refactor full-batch algorithms (e.g., DAEGC, NS4GC) into mini-batch compatible versions. 
By utilizing neighbor sampling and subgraph-based objective approximations, \texttt{PyAGC} decouples GPU memory consumption from graph size. 
This optimization allows complex models to be trained on the massive \texttt{Papers100M} dataset (111M nodes) on a single 32GB GPU, breaking the scalability ceiling of prior benchmarks.

\begin{table*}[t]
    \centering
    \caption{Summary of the benchmark datasets.
    The collection is categorized by scale, spanning five orders of magnitude from Tiny to Massive. 
    The datasets cover a wide range of domains and feature types (Textual, Tabular).
    $\mathcal{H}_e$ and $\mathcal{H}_n$ denote edge homophily and node homophily, respectively.
    }
    \label{tab:datasets}
    \resizebox{\textwidth}{!}{
    \begin{tabular}{lll rr rr c c c c}
        \toprule
        \textbf{Scale} & \textbf{Dataset} & \textbf{Domain} & \textbf{\#Nodes} & \textbf{\#Edges} & \textbf{Avg. Deg.} & \textbf{\#Feat.} & \textbf{Feat. Type} & \textbf{\#Clus.} & \textbf{$\mathcal{H}_e$} & \textbf{$\mathcal{H}_n$} \\
        \midrule
        \multirow{2}{*}{Tiny} 
          & Cora & Citation & 2,708 & 10,556 & 3.9 & 1,433 & Textual & 7 & 0.81 & 0.83 \\
          & Photo & Co-purchase & 7,650 & 238,162 & 31.1 & 745 & Textual & 8 & 0.83 & 0.84 \\
        \midrule
        \multirow{3}{*}{Small} 
          & Physics & Co-author & 34,493 & 495,924 & 14.4 & 8,415 & Textual & 5 & 0.93 & 0.92 \\
          & HM & Co-purchase & 46,563 & 21,461,990 & 460.9 & 120 & \textit{Tabular} & 21 & 0.16 & 0.35 \\
          & Flickr & Social & 89,250 & 899,756 & 10.1 & 500 & Textual & 7 & 0.32 & 0.32 \\
        \midrule
        \multirow{3}{*}{Medium} 
          & ArXiv & Citation & 169,343 & 1,166,243 & 6.9 & 128 & Textual & 40 & 0.65 & 0.64 \\
          & Reddit & Social & 232,965 & 23,213,838 & 99.6 & 602 & Textual & 41 & 0.78 & 0.81 \\
          & MAG & Citation & 736,389 & 10,792,672 & 14.7 & 128 & Textual & 349 & 0.30 & 0.31 \\
        \midrule
        \multirow{3}{*}{Large} 
          & Pokec & Social & 1,632,803 & 44,603,928 & 27.3 & 56 & \textit{Tabular} & 183 & 0.43 & 0.39 \\
          & Products & Co-purchase & 2,449,029 & 61,859,140 & 25.4 & 100 & Textual & 47 & 0.81 & 0.82 \\
          & WebTopic & Web & 2,890,331 & 24,754,822 & 8.6 & 528 & \textit{Tabular} & 28 & 0.22 & 0.18 \\
        \midrule
        Massive & Papers100M$^{\dagger}$ & Citation & 111,059,956 & 1,615,685,872 & 14.5 & 128 & Textual & 172 & 0.57 & 0.50 \\
        \bottomrule
        \multicolumn{11}{p{\textwidth}}{\footnotesize \textit{Note:} For \texttt{Papers100M}$^{\dagger}$, ground-truth labels are available for a subset of $\approx$1.5M arXiv papers. The reported homophily metrics ($\mathcal{H}_e, \mathcal{H}_n$) are calculated based on the induced subgraph of these labeled nodes.}
    \end{tabular}%
    }
\vspace{-0.3cm}
\end{table*}

\section{Benchmark Setup}
\label{sec:setup}

\subsection{Dataset Curation}
\label{sec:dataset}
To provide a rigorous and comprehensive evaluation, we curate a diverse collection of 12 attributed graph datasets, summarized in Table~\ref{tab:datasets}. Our selection criteria move beyond the traditional Cora-centric paradigm, deliberately incorporating large-scale graphs from the Open Graph Benchmark~\cite{OGB} and industrial graphs from the recent GraphLand benchmark~\cite{GraphLand}. 
We provide detailed dataset descriptions in Appendix~\ref{app:datasets}. 
The collection is characterized by four core dimensions of diversity:

\vpara{Scale Diversity.}
Our atlas spans five orders of magnitude.
Beyond canonical \textit{Tiny} benchmarks (\texttt{Cora}, \texttt{Photo}) for sanity checking, the core focuses on \textit{Medium} to \textit{Large} production-scale graphs ($10^5 \le N < 10^7$) such as \texttt{Reddit} and \texttt{WebTopic}, culminating in the \textit{Massive} \texttt{Papers100M} dataset (111M nodes, 1.6B edges) as the ultimate stress test for industrial-grade scalability.

\vpara{Attribute Diversity.}
A critical gap in existing AGC research is over-reliance on textual graphs with clean bag-of-words or sentence-embedding features.
We introduce tabular graphs \texttt{HM}, \texttt{Pokec}, and \texttt{WebTopic}~\cite{GraphLand}, whose skewed, multi-modal, and noise-heavy heterogeneous features reflect the true complexity of industrial node metadata (e.g., user demographics, transaction counts).

\vpara{Structural Diversity.}
Our atlas spans the full homophily-heterophily spectrum: from high-homophily academic graphs (e.g., \texttt{Physics}, $\mathcal{H}_e{=}0.93$) to heterophilous industrial graphs (\texttt{HM}, \texttt{WebTopic}, $\mathcal{H}_e{<}0.25$). Structural density also varies widely, from sparse citation graphs (\texttt{ArXiv}, Avg. Deg. 6.9) to dense co-purchase networks (\texttt{HM}, Avg. Deg. 460.9), ensuring AGC models are tested on both over-smoothing-prone and noisy heterophilous topologies.

\vpara{Domain Diversity.}
By integrating industrial data, we extend evaluation from narrow \textit{Citation} and \textit{Co-author} networks to \textit{E-commerce} (\texttt{Products}, \texttt{HM}), \textit{Social Networks} (\texttt{Reddit}, \texttt{Pokec}, \texttt{Flickr}), and the \textit{Web Graph} (\texttt{WebTopic}), ensuring that discovered insights generalize across the broader ecosystem of graph machine learning applications where ground-truth clusters are intrinsically tied to specific industrial behaviors like fraud, co-purchasing, or user navigation.

\begin{table*}[t]
  \centering
  \caption{A taxonomy of representative Attributed Graph Clustering methods under the Encode-Cluster-Optimize framework. 
  $\mathcal{E}$: Encoder Type (Parametric vs. Non-Parametric); $\mathcal{C}$: Clusterer Type (Differentiable vs. Discrete); $\mathcal{O}$: Optimization Strategy (Joint vs. Decoupled).  \textbf{Complexity} denotes the time/space complexity, where $N=|\mathcal{V}|$ and $M=|\mathcal{E}|$. $O(N+M)$ indicates linear scalability on sparse graphs, while $O(N^2)$ indicates quadratic bottlenecks.}
  \label{tab:eco_taxonomy}
  \resizebox{\textwidth}{!}{
  \begin{tabular}{l l lllll}
      \toprule
      \textbf{Method} & \textbf{Venue} & \textbf{Encode ($\mathcal{E}$)} & \textbf{Cluster ($\mathcal{C}$)} & \textbf{Optimize ($\mathcal{O}$)} & \textbf{Core Objective} & \textbf{Complexity} \\
      \midrule
      \multicolumn{7}{c}{\textit{\textbf{Non-Parametric \& Decoupled Methods}}} \\
      \midrule
      \textbf{AGC}~\cite{AGC} & IJCAI'19 & Fixed Filter & Discrete (Spectral) & Decoupled & Adaptive Smoothing & $O(N^2)$ \\
      \textbf{SSGC}~\cite{SSGC} & ICLR'21 & Adaptive Filter & Discrete (KMeans) & Decoupled & Markov Diffusion & $O(N+M)$ \\
      \textbf{NAFS}~\cite{NAFS} & ICML'22 & Adaptive Filter & Discrete (KMeans) & Decoupled & Smoothing + Ensemble & $O(N+M)$ \\
      \textbf{SAGSC}~\cite{SAGSC} & AAAI'23 & Fixed Filter & Discrete (Subspace) & Decoupled & Self-Expressive Subspace & $O(N+M)$ \\
      \textbf{MS2CAG}~\cite{MS2CAG} & KDD'25 & Fixed Filter & Discrete (SNEM~\cite{SNEM}) & Decoupled & Rank-Constrained SVD & $O(N+M)$ \\
      \midrule
      \multicolumn{7}{c}{\textit{\textbf{Deep Decoupled Methods (Parametric Encoder + Post-hoc Clustering)}}} \\
      \midrule
      \textbf{GAE}~\cite{GAE} & NIPS-W'16 & Parametric (GCN) & Discrete (KMeans) & Decoupled & Graph Reconstruction & $O(N^2)$ \\
      \textbf{DGI}~\cite{DGI} & ICLR'19 & Parametric (GCN) & Discrete (KMeans) & Decoupled & Mutual Info Maximization & $O(N+M)$ \\
      \textbf{CCASSG}~\cite{CCASSG} & NeurIPS'21 & Parametric (GCN) & Discrete (KMeans) & Decoupled & Redundancy Reduction & $O(N+M)$ \\
      \textbf{S3GC}~\cite{S3GC} & NeurIPS'22 & Parametric (GCN) & Discrete (KMeans) & Decoupled & Contrastive (Random Walk) & $O(N^2)$ \\
      \textbf{NS4GC}~\cite{NS4GC} & TKDE'24 & Parametric (GCN) & Discrete (KMeans) & Decoupled & Contrastive (Node Similarity) & $O(N^2)$ \\
      \textbf{MAGI}~\cite{MAGI} & KDD'24 & Parametric (GNN) & Discrete (KMeans) & Decoupled & Contrastive (Modularity) & $O(N^2)$ \\
      \midrule
      \multicolumn{7}{c}{\textit{\textbf{Deep Joint Methods (Parametric Encoder + Differentiable Clustering)}}} \\
      \midrule
      \textbf{DAEGC}~\cite{DAEGC} & IJCAI'19 & Parametric (GAT) & Differen. (Prototype) & Joint & Reconstruction + KL Div. & $O(N^2)$ \\
      \textbf{DinkNet}~\cite{DinkNet} & ICML'23 & Parametric (GCN) & Differen. (Prototype) & Joint & Dilation + Shrink Loss & $O(N+M)$ \\
      \textbf{MinCut}~\cite{MinCut} & ICML'20 & Parametric (GCN) & Differen. (Softmax) & Joint & Cut Minimization & $O(N+M)$ \\
      \textbf{DMoN}~\cite{DMoN} & JMLR'23 & Parametric (GCN) & Differen. (Softmax) & Joint & Modularity Maximization & $O(N+M)$ \\
      \textbf{Neuromap}~\cite{Neuromap} & NeurIPS'24 & Parametric (GCN) & Differen. (Softmax) & Joint & Map Equation & $O(N+M)$ \\
      \bottomrule
  \end{tabular}
  }
\end{table*}

\subsection{Evaluated Representative Algorithms}
\label{sec:baselines}

We evaluate a comprehensive suite of 17 representative AGC methods, selected to cover the entire spectrum of the ECO framework: 
1) \emph{Traditional Methods:} attribute-only KMeans~\cite{KMeans} and structure-only Node2Vec~\cite{Node2Vec}; 
2) \emph{Non-Parametric Methods:} SSGC~\cite{SSGC}, SAGSC~\cite{SAGSC} and MS2CAG~\cite{MS2CAG}; 
3) \emph{Deep Decoupled Methods:} GAE~\cite{GAE}, DGI~\cite{DGI}, CCASSG~\cite{CCASSG}, S3GC~\cite{S3GC}, NS4GC~\cite{NS4GC} and MAGI~\cite{MAGI}; 
4) \emph{Deep Joint Methods:} DAEGC~\cite{DAEGC}, DinkNet~\cite{DinkNet}, MinCut~\cite{MinCut}, DMoN~\cite{DMoN}, and Neuromap~\cite{Neuromap}.
Table~\ref{tab:eco_taxonomy} provides a detailed classification of these methods within the ECO framework, while more detailed descriptions are presented in Appendix~\ref{app:baselines}.

\subsection{Holistic Evaluation Protocol}
\label{sec:evaluation_protocol}

To rectify the \textit{supervised metric paradox} and \textit{scalability bottleneck}, we institute a multi-view evaluation protocol comprising supervised alignment, unsupervised quality, and computational efficiency:

\vpara{Supervised Alignment Metrics.}
To maintain backward compatibility with existing literature, we report \textbf{Accuracy (ACC)}, \textbf{Macro-F1 Score (F1)}, \textbf{Normalized Mutual Information (NMI)}, and \textbf{Adjusted Rand Index (ARI)}. 
These metrics measure how well the discovered clusters align with human-annotated labels. 
However, we explicitly warn that on industrial datasets (e.g., \texttt{WebTopic}), ground truth labels often represent only one specific view (e.g., website topic) of a multi-faceted entity, and low ACC does not necessarily imply poor structural clustering.

\vpara{Unsupervised Structural Metrics.}
Recognizing that real-world clustering is often label-free, we mandate the reporting of structural metrics to assess the intrinsic quality of the partition $\mathcal{C} = \{C_i\}_{i=1}^K$. 
These metrics evaluate the community structure based purely on topological connectivity:
\begin{itemize}[leftmargin=*]
    \item \textbf{Modularity ($\mathcal{Q}$)}~\cite{Modularity}: Measures the strength of the division of a network into communities. It compares the density of edges within clusters to the expected density in a random graph with the same degree distribution:
    \begin{equation}
        \mathcal{Q} = \frac{1}{2|\mathcal{E}|} \sum_{i, j} \left( \boldsymbol{A}_{ij} - \frac{d_i d_j}{2|\mathcal{E}|} \right) \delta(c_i, c_j)
    \end{equation}
    where $|\mathcal{E}|$ is the total number of edges, $d_i$ is the degree of node $v_i$, $c_i$ is the cluster assignment of node $v_i$, and $\delta(c_i, c_j) = 1$ if $v_i$ and $v_j$ are in the same cluster.
    Higher modularity indicates stronger community structure.

    \item \textbf{Conductance ($\mathcal{K}$)}~\cite{Conductance}: Measures the separability of clusters by quantifying the fraction of total edge volume that points outside the cluster. We report the average conductance over all $K$ clusters:
    \begin{equation}
        \mathcal{K} = \frac{1}{K} \sum_{k=1}^{K} \frac{c_{C_k}}{2m_{C_k} + c_{C_k}}
    \end{equation}
    where for a given cluster $C_k$, $c_{C_k} = |\{(u, v) \in \mathcal{E} : u \in C_k, v \notin C_k\}|$ represents the number of edges on the boundary (cut size), and $m_{C_k} = |\{(u, v) \in \mathcal{E} : u \in C_k, v \in C_k\}|$ represents the number of internal edges. The denominator $2m_{C_k} + c_{C_k}$ effectively represents the total volume (sum of degrees) of nodes in set $C_k$. Lower conductance implies better isolation of communities.
\end{itemize}

\vpara{Efficiency and Scalability Profiling.}
We mandate the reporting of the total training and clustering time and peak GPU memory consumption to provide practitioners with a clear understanding of the trade-offs between model complexity and operational cost.

\subsection{Implementation Details}
Experiments are conducted on a single NVIDIA V100 (32GB) GPU to emphasize scalability. 
All deep models are optimized with Adam and evaluated over 5 random runs. 
All hyperparameters are managed via configuration files to ensure reproducibility. 
Please kindly see Appendix~\ref{app:implementation_details} for more details and see \href{https://github.com/Cloudy1225/PyAGC}{\underline{\texttt{GitHub}}} for the code, documentation, and specific hyperparameters.

\begin{table*}[t]
\centering
\caption{
Clustering performance comparison measured by NMI and ACC ($\%$) (Mean $\pm$ SD). The \colorbox{colorBest}{\textbf{best}} and \colorbox{colorSecond}{second-best} results are highlighted. "--" denotes OOM errors as these methods strictly require full-graph processing.}
\label{tab:nmi_acc}
\begin{adjustbox}{width=\textwidth, center}
\begin{tabular}{llll cccccccccc cc}
\toprule
~ & \multirow{2}{*}{\textbf{Model}} & \multirow{2}{*}{\textbf{Metric}} & \multicolumn{2}{c}{\textbf{Tiny}} & \multicolumn{3}{c}{\textbf{Small}} & \multicolumn{3}{c}{\textbf{Medium}} & \multicolumn{3}{c}{\textbf{Large}} & \multicolumn{1}{c}{\textbf{Massive}} \\
\cmidrule(lr){4-5} \cmidrule(lr){6-8} \cmidrule(lr){9-11} \cmidrule(lr){12-14} \cmidrule(lr){15-15}
& & & \textbf{Cora} & \textbf{Photo} & \textbf{Physics} & \textbf{HM} & \textbf{Flickr} & \textbf{ArXiv} & \textbf{Reddit} & \textbf{MAG} & \textbf{Pokec} & \textbf{Products} & \textbf{WebTop.} & \textbf{Papers.} \\
\midrule

\multirow{4}{*}{\rotatebox{90}{Traditional}} 
& \multirow{2}{*}{KMeans}   & NMI & \res{13.89}{4.46} & \res{31.70}{0.49} & \res{52.00}{0.13} & \res{10.17}{0.11} & \res{1.21}{0.07} & \res{22.59}{0.03} & \res{11.08}{0.14} & \res{28.32}{0.06} & \res{1.41}{0.00} & \res{29.24}{0.11} & \res{2.45}{0.05} & \res{37.77}{0.07} \\
&                          & ACC & \res{33.09}{4.00} & \res{42.54}{2.54} & \res{53.41}{0.13} & \res{14.12}{0.25} & \res{26.02}{1.73} & \res{16.95}{0.38} & \res{8.59}{0.06} & \res{7.09}{0.03} & \res{1.91}{0.01} & \res{21.23}{0.74} & \res{9.63}{0.34} & \res{15.16}{0.21} \\
\cmidrule{2-15}
& \multirow{2}{*}{Node2Vec} & NMI & \res{44.95}{1.49} & \res{66.02}{1.05} & \res{54.94}{0.01} & \res{7.60}{0.13} & \res{5.59}{0.00} & \res{38.93}{0.15} & \res{79.23}{0.38} & \res{37.73}{0.03} & \second{31.92}{0.02} & \res{50.97}{0.20} & \res{5.80}{0.09} & \res{51.29}{0.05} \\
&                          & ACC & \res{60.52}{2.81} & \res{73.08}{1.06} & \res{57.55}{0.01} & \res{13.28}{0.41} & \res{24.44}{0.01} & \res{25.97}{0.57} & \res{70.67}{1.68} & \res{10.86}{0.09} & \second{16.62}{0.19} & \res{36.59}{0.74} & \res{10.71}{0.12} & \best{25.31}{0.26} \\
\midrule

\multirow{6}{*}{\rotatebox{90}{Non-Parametric}} 
& \multirow{2}{*}{SSGC}     & NMI & \res{51.85}{1.05} & \res{70.75}{1.39} & \res{64.49}{4.20} & \res{12.15}{0.11} & \res{4.48}{0.19} & \res{46.12}{0.16} & \res{51.61}{0.36} & \second{41.52}{0.04} & \res{4.33}{0.01} & \res{52.04}{0.17} & \res{3.92}{0.11} & -- \\
&                          & ACC & \res{65.38}{1.47} & \res{76.34}{4.04} & \res{65.04}{12.1} & \res{15.57}{0.07} & \res{25.95}{0.42} & \second{38.58}{0.56} & \res{40.00}{1.79} & \res{11.89}{0.10} & \res{2.93}{0.04} & \res{37.02}{0.70} & \res{11.34}{0.18} & -- \\
\cmidrule{2-15}
& \multirow{2}{*}{SAGSC}    & NMI & \res{44.35}{0.06} & \res{58.40}{0.02} & \res{55.95}{0.04} & \res{12.47}{0.13} & \res{6.86}{0.00} & \res{43.27}{0.19} & \second{80.02}{0.38} & \res{40.15}{0.03} & \best{38.33}{0.03} & \res{51.78}{0.24} & \res{9.11}{0.26} & -- \\
&                          & ACC & \res{63.45}{0.03} & \res{67.91}{0.01} & \res{60.15}{0.01} & \res{17.83}{0.11} & \res{26.72}{0.01} & \res{30.87}{0.83} & \second{72.79}{1.57} & \res{11.90}{0.16} & \best{19.86}{0.23} & \res{38.11}{0.69} & \res{13.66}{0.53} & -- \\
\cmidrule{2-15}
& \multirow{2}{*}{MS2CAG}   & NMI & \res{53.64}{0.64} & \second{72.49}{0.79} & \second{72.45}{0.03} & \res{9.75}{0.36} & \res{7.32}{0.02} & \res{44.20}{0.18} & \res{72.85}{0.53} & \res{40.39}{0.07} & \res{2.77}{0.05} & \res{50.71}{0.31} & \res{7.77}{0.08} & -- \\
&                          & ACC & \res{70.29}{0.66} & \second{79.13}{0.05} & \best{90.45}{0.04} & \res{15.06}{0.28} & \res{28.29}{0.06} & \res{35.85}{0.21} & \res{60.40}{2.40} & \res{11.70}{0.10} & \res{2.27}{0.09} & \res{35.41}{0.62} & \res{12.38}{0.12} & -- \\
\midrule

\multirow{12}{*}{\rotatebox{90}{Deep Decoupled}} 
& \multirow{2}{*}{GAE}      & NMI & \res{50.09}{0.07} & \res{61.36}{0.14} & \res{68.15}{0.03} & \second{13.58}{0.12} & \res{4.07}{0.05} & \res{40.86}{0.15} & \res{45.90}{0.27} & \res{39.30}{0.03} & \res{3.79}{0.04} & \res{42.55}{0.09} & \res{3.89}{0.14} & \res{42.84}{0.02} \\
&                          & ACC & \res{67.53}{0.13} & \res{71.54}{0.08} & \res{84.99}{0.05} & \res{16.58}{0.32} & \best{38.42}{1.07} & \res{24.39}{0.31} & \res{34.56}{0.81} & \res{11.21}{0.07} & \res{2.49}{0.02} & \res{27.70}{0.43} & \res{11.86}{0.06} & \res{15.71}{0.08} \\
\cmidrule{2-15}
& \multirow{2}{*}{DGI}      & NMI & \res{56.22}{0.87} & \res{67.77}{0.54} & \res{74.39}{0.35} & \res{11.67}{0.06} & \res{6.88}{0.01} & \res{42.29}{0.08} & \res{72.93}{0.14} & \res{39.39}{0.03} & \res{4.20}{0.01} & \res{41.28}{0.13} & \res{5.91}{0.02} & \res{49.28}{0.07} \\
&                          & ACC & \res{72.22}{2.92} & \res{77.57}{0.28} & \res{89.87}{0.32} & \res{15.78}{0.32} & \res{24.92}{0.17} & \res{30.49}{0.64} & \res{62.52}{0.34} & \res{10.60}{0.11} & \res{2.23}{0.02} & \res{28.99}{0.52} & \res{11.69}{0.10} & \res{20.70}{0.14} \\
\cmidrule{2-15}
& \multirow{2}{*}{CCASSG}   & NMI & \res{58.74}{0.87} & \res{64.54}{3.23} & \res{70.94}{0.04} & \res{11.93}{0.15} & \res{4.67}{0.01} & \res{44.69}{0.04} & \res{49.63}{0.07} & \res{40.40}{0.02} & \res{1.35}{0.01} & \res{50.89}{0.31} & \res{3.79}{0.18} & \best{53.82}{0.05} \\
&                          & ACC & \res{73.35}{1.92} & \res{70.90}{2.27} & \res{87.57}{0.05} & \res{15.78}{0.14} & \res{23.03}{0.35} & \res{30.68}{0.27} & \res{37.91}{0.49} & \res{11.90}{0.05} & \res{1.95}{0.01} & \res{37.99}{0.42} & \res{11.65}{0.91} & \second{25.18}{0.14} \\
\cmidrule{2-15}
& \multirow{2}{*}{S3GC}     & NMI & \res{55.45}{1.12} & \res{68.27}{2.22} & \res{70.87}{0.05} & \res{11.57}{0.08} & \second{7.84}{0.00} & \second{47.11}{0.12} & \best{83.45}{0.39} & \res{39.80}{0.03} & \res{6.04}{0.00} & \second{53.43}{0.16} & \res{7.83}{0.02} & \res{44.18}{0.04} \\
&                          & ACC & \res{70.33}{1.98} & \res{75.92}{3.26} & \res{87.30}{0.05} & \res{16.49}{0.22} & \res{28.48}{0.00} & \res{35.57}{0.60} & \best{81.86}{1.59} & \res{12.15}{0.05} & \res{2.97}{0.03} & \second{40.45}{0.62} & \res{12.75}{0.07} & \res{17.98}{0.19} \\
\cmidrule{2-15}
& \multirow{2}{*}{NS4GC}    & NMI & \best{59.40}{0.48} & \best{72.62}{0.79} & \best{75.38}{0.08} & \best{15.28}{0.17} & \res{6.19}{0.01} & \best{48.39}{0.24} & \res{56.71}{0.05} & \best{41.64}{0.01} & \res{7.19}{0.04} & \best{54.63}{0.14} & \best{10.05}{0.04} & \res{49.83}{0.04} \\
&                          & ACC & \best{74.92}{0.59} & \best{79.21}{0.03} & \second{90.15}{0.07} & \second{18.40}{0.42} & \res{24.27}{0.06} & \best{39.01}{1.07} & \res{43.80}{0.52} & \second{13.28}{0.09} & \res{3.19}{0.01} & \best{41.83}{0.40} & \res{15.69}{0.14} & \res{22.52}{0.16} \\
\cmidrule{2-15}
& \multirow{2}{*}{MAGI}     & NMI & \second{58.94}{0.41} & \res{68.65}{0.16} & \res{66.08}{0.13} & \res{11.24}{0.46} & \res{6.31}{0.16} & \res{46.53}{0.16} & \res{72.53}{0.19} & \res{41.34}{0.02} & \res{8.76}{0.02} & \res{44.58}{0.14} & \second{9.27}{0.20} & \second{53.06}{0.04} \\
&                          & ACC & \second{74.22}{0.57} & \res{76.66}{0.18} & \res{62.82}{0.72} & \best{18.49}{0.83} & \res{24.28}{0.46} & \res{37.32}{0.74} & \res{64.78}{1.38} & \res{13.25}{0.07} & \res{3.90}{0.03} & \res{33.15}{0.34} & \res{14.37}{0.76} & \res{24.61}{0.13} \\
\midrule

\multirow{10}{*}{\rotatebox{90}{Deep Joint}} 
& \multirow{2}{*}{DAEGC}    & NMI & \res{46.88}{1.71} & \res{63.60}{0.02} & \res{57.04}{0.03} & \res{11.46}{0.09} & \res{4.23}{0.03} & \res{39.81}{0.29} & \res{40.84}{0.76} & \res{29.00}{2.17} & \res{3.85}{0.21} & \res{9.18}{3.05} & \res{4.79}{0.29} & \res{28.59}{0.06} \\
&                          & ACC & \res{64.06}{3.18} & \res{77.78}{0.01} & \res{58.11}{0.06} & \res{17.37}{0.31} & \res{28.51}{0.35} & \res{28.68}{0.64} & \res{27.57}{0.33} & \res{10.83}{0.51} & \res{3.99}{0.03} & \res{14.83}{2.09} & \second{16.42}{0.61} & \res{16.66}{0.35} \\
\cmidrule{2-15}
& \multirow{2}{*}{DinkNet}  & NMI & \res{55.56}{0.18} & \res{64.74}{0.06} & \res{57.34}{0.06} & \res{11.26}{0.06} & \res{6.67}{0.05} & \res{37.24}{0.57} & \res{54.95}{0.24} & \res{37.12}{0.05} & \res{3.91}{0.01} & \res{38.43}{0.18} & \res{5.47}{0.04} & \res{45.86}{0.04} \\
&                          & ACC & \res{72.49}{0.50} & \res{74.03}{0.25} & \res{55.67}{0.26} & \res{15.79}{0.28} & \second{29.94}{0.38} & \res{25.47}{0.88} & \res{38.51}{0.82} & \res{10.14}{0.08} & \res{2.73}{0.04} & \res{26.78}{0.37} & \res{11.52}{0.55} & \res{18.63}{0.25} \\
\cmidrule{2-15}
& \multirow{2}{*}{MinCut}   & NMI & \res{40.80}{1.83} & \res{62.34}{2.39} & \res{56.94}{2.06} & \res{7.42}{0.37} & \res{7.59}{0.19} & \res{39.00}{1.02} & \res{48.85}{3.58} & \res{38.21}{0.12} & \res{5.78}{0.18} & \res{35.80}{0.71} & \res{6.07}{0.52} & -- \\
&                          & ACC & \res{52.95}{6.65} & \res{69.46}{2.22} & \res{54.96}{2.46} & \res{13.87}{0.43} & \res{27.64}{1.87} & \res{27.88}{1.20} & \res{34.09}{2.08} & \res{11.60}{0.32} & \res{3.17}{0.09} & \res{24.88}{0.74} & \res{11.67}{0.95} & -- \\
\cmidrule{2-15}
& \multirow{2}{*}{DMoN}     & NMI & \res{43.84}{2.29} & \res{62.74}{2.67} & \res{58.30}{0.37} & \res{7.59}{0.92} & \best{7.89}{0.09} & \res{38.77}{0.28} & \res{50.66}{0.38} & \res{38.30}{0.12} & \res{8.42}{0.08} & \res{34.80}{0.85} & \res{7.49}{0.37} & -- \\
&                          & ACC & \res{52.99}{4.70} & \res{72.68}{1.87} & \res{57.87}{0.34} & \res{12.32}{0.58} & \res{28.64}{0.69} & \res{25.89}{1.12} & \res{39.29}{1.29} & \res{10.81}{0.08} & \res{4.40}{0.07} & \res{25.40}{1.37} & \res{12.70}{0.73} & -- \\
\cmidrule{2-15}
& \multirow{2}{*}{Neuromap} & NMI & \res{46.98}{2.61} & \res{61.84}{2.34} & \res{56.53}{1.63} & \res{7.44}{0.17} & \res{7.77}{0.57} & \res{40.70}{0.80} & \res{47.81}{0.28} & \res{39.14}{0.20} & \res{8.53}{0.25} & \res{34.80}{1.46} & \res{7.19}{0.51} & -- \\
&                          & ACC & \res{56.96}{3.98} & \res{71.89}{4.19} & \res{55.00}{2.32} & \res{12.73}{0.25} & \res{28.86}{2.39} & \res{33.31}{1.07} & \res{32.64}{0.20} & \best{15.95}{0.32} & \res{5.12}{0.21} & \res{31.87}{1.93} & \best{16.82}{1.69} & -- \\
\bottomrule
\end{tabular}
\end{adjustbox}
\vspace{-0.3cm}
\end{table*}

\section{Benchmark Results}
\label{sec:results}

In this section, we present a comprehensive analysis of the 17 evaluated methods across our 12 datasets. We aim to answer three critical questions: 
(1) How well do current AGC methods generalize from academic datasets to complex industrial graphs? 
(2) Do supervised metrics (e.g., NMI) accurately reflect intrinsic clustering quality in label-scarce settings? 
(3) Can the proposed \texttt{PyAGC} library effectively scale deep clustering algorithms to massive graphs?

\begin{table*}[t]
\centering
\caption{
Clustering performance comparison measured by ARI and F1 ($\%$) (Mean $\pm$ SD). The \colorbox{colorBest}{\textbf{best}} and \colorbox{colorSecond}{second-best} results are highlighted. "--" denotes OOM errors as these methods strictly require full-graph processing.}
\label{tab:ari_f1}
\begin{adjustbox}{width=\textwidth, center}
\begin{tabular}{llll cccccccccc cc}
\toprule
~ & \multirow{2}{*}{\textbf{Model}} & \multirow{2}{*}{\textbf{Metric}} & \multicolumn{2}{c}{\textbf{Tiny}} & \multicolumn{3}{c}{\textbf{Small}} & \multicolumn{3}{c}{\textbf{Medium}} & \multicolumn{3}{c}{\textbf{Large}} & \multicolumn{1}{c}{\textbf{Massive}} \\
\cmidrule(lr){4-5} \cmidrule(lr){6-8} \cmidrule(lr){9-11} \cmidrule(lr){12-14} \cmidrule(lr){15-15}
& & & \textbf{Cora} & \textbf{Photo} & \textbf{Physics} & \textbf{HM} & \textbf{Flickr} & \textbf{ArXiv} & \textbf{Reddit} & \textbf{MAG} & \textbf{Pokec} & \textbf{Products} & \textbf{WebTop.} & \textbf{Papers.} \\
\midrule

\multirow{4}{*}{\rotatebox{90}{Traditional}} 
& \multirow{2}{*}{KMeans}   & ARI & \res{6.95}{2.61} & \res{20.01}{1.01} & \res{32.34}{0.29} & \res{2.96}{0.09} & \res{1.06}{0.15} & \res{7.23}{0.15} & \res{2.93}{0.05} & \res{2.77}{0.03} & \res{0.23}{0.00} & \res{9.22}{0.33} & \res{0.87}{0.04} & \res{7.63}{0.08} \\
&                          & F1  & \res{25.99}{2.91} & \res{42.68}{3.32} & \res{52.12}{0.11} & \res{12.50}{0.20} & \res{15.16}{0.85} & \res{12.97}{0.18} & \res{7.25}{0.08} & \res{5.71}{0.07} & \res{0.98}{0.01} & \res{12.77}{0.41} & \res{4.64}{0.04} & \res{11.02}{0.19} \\
\cmidrule{2-15}
& \multirow{2}{*}{Node2Vec} & ARI & \res{33.61}{2.49} & \res{57.11}{1.39} & \res{39.86}{0.01} & \res{1.84}{0.07} & \res{1.52}{0.00} & \res{16.69}{0.73} & \res{64.34}{1.64} & \res{4.75}{0.07} & \res{9.27}{0.08} & \res{18.53}{1.05} & \res{1.53}{0.03} & \res{16.17}{0.15} \\
&                          & F1  & \res{61.66}{2.52} & \res{69.33}{1.18} & \res{57.91}{0.02} & \res{11.65}{0.46} & \res{19.68}{0.01} & \res{20.96}{0.29} & \res{68.61}{2.12} & \res{8.90}{0.09} & \res{12.66}{0.15} & \res{25.41}{0.24} & \res{7.40}{0.16} & \best{19.38}{0.19} \\
\midrule

\multirow{6}{*}{\rotatebox{90}{Non-Parametric}} 
& \multirow{2}{*}{SSGC}     & ARI & \res{41.25}{1.69} & \res{59.17}{3.81} & \res{56.50}{13.07} & \res{3.18}{0.02} & \res{3.29}{0.11} & \best{33.04}{0.29} & \res{33.61}{1.29} & \res{5.84}{0.09} & \res{0.42}{0.01} & \res{18.84}{0.40} & \res{1.75}{0.04} & -- \\
&                          & F1  & \res{65.16}{1.32} & \res{70.17}{3.78} & \res{63.57}{10.46} & \res{14.84}{0.13} & \res{18.64}{0.48} & \second{25.44}{0.59} & \res{29.15}{1.33} & \second{9.78}{0.05} & \res{1.60}{0.03} & \res{24.01}{0.71} & \res{5.66}{0.04} & -- \\
\cmidrule{2-15}
& \multirow{2}{*}{SAGSC}    & ARI & \res{38.83}{0.02} & \res{43.66}{0.01} & \res{46.37}{0.02} & \res{4.12}{0.08} & \res{3.27}{0.00} & \res{20.69}{0.69} & \second{65.93}{1.44} & \res{4.89}{0.11} & \second{11.58}{0.18} & \res{19.36}{0.29} & \res{2.91}{0.10} & -- \\
&                          & F1  & \res{61.03}{0.02} & \res{69.40}{0.01} & \res{60.87}{0.02} & \second{16.08}{0.10} & \res{20.97}{0.01} & \res{25.37}{0.85} & \second{70.43}{1.60} & \res{9.67}{0.20} & \second{16.12}{0.26} & \second{25.56}{0.53} & \second{9.00}{0.30} & -- \\
\cmidrule{2-15}
& \multirow{2}{*}{MS2CAG}   & ARI & \res{45.76}{0.98} & \best{63.88}{0.54} & \res{80.89}{0.05} & \res{3.50}{0.11} & \res{3.84}{0.03} & \res{28.84}{0.18} & \res{56.56}{2.89} & \res{5.16}{0.09} & \res{0.42}{0.03} & \res{17.97}{0.65} & \res{2.36}{0.03} & -- \\
&                          & F1  & \res{69.17}{0.65} & \second{74.94}{0.95} & \best{87.60}{0.05} & \res{12.95}{0.27} & \res{21.81}{0.03} & \res{24.58}{0.50} & \res{55.37}{1.78} & \best{9.97}{0.14} & \res{1.51}{0.02} & \res{24.67}{0.31} & \res{8.52}{0.07} & -- \\
\midrule

\multirow{12}{*}{\rotatebox{90}{Deep Decoupled}} 
& \multirow{2}{*}{GAE}      & ARI & \res{45.63}{0.15} & \res{52.39}{0.12} & \res{74.25}{0.11} & \second{4.21}{0.23} & \best{5.53}{0.30} & \res{15.95}{0.40} & \res{27.88}{0.46} & \res{5.34}{0.08} & \res{0.34}{0.01} & \res{14.49}{0.25} & \res{1.98}{0.06} & \res{9.80}{0.06} \\
&                          & F1  & \res{64.48}{0.11} & \res{67.52}{0.09} & \res{80.04}{0.04} & \res{15.20}{0.22} & \res{18.83}{0.13} & \res{18.57}{0.22} & \res{26.89}{0.50} & \res{8.97}{0.09} & \res{1.53}{0.02} & \res{18.35}{0.47} & \res{5.81}{0.14} & \res{11.84}{0.10} \\
\cmidrule{2-15}
& \multirow{2}{*}{DGI}      & ARI & \res{52.03}{3.22} & \res{58.38}{0.46} & \second{83.05}{0.20} & \res{3.11}{0.06} & \res{4.98}{0.01} & \res{22.77}{0.28} & \res{59.28}{0.54} & \res{5.55}{0.07} & \res{0.32}{0.00} & \res{14.67}{0.21} & \res{2.18}{0.06} & \res{12.17}{0.16} \\
&                          & F1  & \res{70.07}{2.56} & \res{72.77}{1.71} & \second{85.36}{0.58} & \res{14.20}{0.23} & \res{19.92}{0.15} & \res{21.47}{0.30} & \res{51.97}{0.79} & \res{8.56}{0.05} & \res{1.59}{0.01} & \res{17.87}{0.32} & \res{7.10}{0.06} & \res{15.53}{0.25} \\
\cmidrule{2-15}
& \multirow{2}{*}{CCASSG}   & ARI & \res{51.84}{3.32} & \res{51.80}{4.28} & \res{79.39}{0.11} & \res{3.20}{0.05} & \res{3.34}{0.03} & \res{19.18}{0.18} & \res{31.28}{0.32} & \res{6.11}{0.05} & \res{0.23}{0.00} & \res{22.43}{0.70} & \res{1.40}{0.29} & \best{19.02}{0.17} \\
&                          & F1  & \res{72.29}{1.48} & \res{69.69}{1.95} & \res{82.45}{0.07} & \res{14.29}{0.24} & \res{16.67}{0.26} & \res{23.78}{0.25} & \res{30.09}{0.32} & \res{8.56}{0.06} & \res{0.98}{0.01} & \res{22.83}{0.39} & \res{5.17}{0.16} & \second{17.41}{0.12} \\
\cmidrule{2-15}
& \multirow{2}{*}{S3GC}     & ARI & \res{48.12}{2.56} & \res{59.09}{2.86} & \res{79.38}{0.08} & \res{3.30}{0.05} & \second{5.45}{0.00} & \res{27.08}{0.55} & \best{83.27}{2.36} & \second{7.51}{0.11} & \res{0.67}{0.01} & \second{23.57}{0.28} & \res{2.56}{0.04} & \res{13.01}{0.16} \\
&                          & F1  & \res{69.04}{1.57} & \res{70.21}{2.55} & \res{81.82}{0.05} & \res{14.59}{0.29} & \best{23.21}{0.00} & \res{24.09}{0.70} & \best{71.35}{1.60} & \res{7.91}{0.05} & \res{2.02}{0.01} & \res{25.19}{0.41} & \res{8.07}{0.02} & \res{11.83}{0.12} \\
\cmidrule{2-15}
& \multirow{2}{*}{NS4GC}    & ARI & \best{56.63}{1.20} & \second{62.64}{0.62} & \best{85.15}{0.06} & \best{4.72}{0.10} & \res{3.37}{0.01} & \second{31.84}{2.30} & \res{39.12}{0.19} & \res{6.85}{0.06} & \res{0.74}{0.01} & \best{23.82}{0.23} & \best{3.86}{0.05} & \res{16.58}{0.20} \\
&                          & F1  & \best{72.81}{0.38} & \res{73.73}{1.11} & \res{84.93}{0.12} & \best{17.23}{0.28} & \res{19.44}{0.03} & \best{26.67}{0.57} & \res{34.95}{0.63} & \res{9.14}{0.08} & \res{2.33}{0.02} & \best{26.52}{0.28} & \res{8.80}{0.15} & \res{15.36}{0.09} \\
\cmidrule{2-15}
& \multirow{2}{*}{MAGI}     & ARI & \second{53.51}{1.09} & \res{58.53}{0.32} & \res{49.76}{0.83} & \res{2.46}{0.27} & \res{3.16}{0.27} & \res{30.38}{1.33} & \res{62.27}{1.24} & \res{6.44}{0.06} & \res{1.06}{0.01} & \res{16.67}{0.37} & \res{2.92}{0.29} & \second{18.25}{0.26} \\
&                          & F1  & \second{72.65}{0.36} & \res{70.82}{0.15} & \res{64.08}{0.22} & \res{14.33}{0.78} & \res{20.81}{0.36} & \res{25.17}{0.59} & \res{55.26}{1.62} & \res{9.46}{0.06} & \second{2.90}{0.02} & \res{21.12}{0.19} & \best{9.53}{0.27} & \res{17.10}{0.23} \\
\midrule

\multirow{10}{*}{\rotatebox{90}{Deep Joint}} 
& \multirow{2}{*}{DAEGC}    & ARI & \res{38.36}{2.67} & \res{56.37}{0.02} & \res{37.88}{0.05} & \res{3.47}{0.07} & \res{4.61}{0.10} & \res{17.53}{0.40} & \res{13.43}{1.24} & \res{4.29}{0.36} & \res{0.61}{0.04} & \res{2.35}{0.90} & \res{2.58}{0.38} & \res{8.68}{1.39} \\
&                          & F1  & \res{64.54}{2.32} & \best{76.26}{0.01} & \res{54.39}{0.05} & \res{14.73}{0.05} & \res{19.24}{0.09} & \res{20.64}{0.28} & \res{8.23}{0.93} & \res{7.25}{0.43} & \res{1.73}{0.07} & \res{8.02}{1.81} & \res{7.34}{0.19} & \res{4.08}{0.26} \\
\cmidrule{2-15}
& \multirow{2}{*}{DinkNet}  & ARI & \res{52.45}{0.41} & \res{53.54}{0.23} & \res{40.79}{0.23} & \res{2.96}{0.03} & \res{1.56}{0.33} & \res{14.93}{0.50} & \res{33.70}{1.24} & \res{5.10}{0.04} & \res{0.29}{0.02} & \res{13.67}{0.40} & \res{1.86}{0.25} & \res{14.21}{0.19} \\
&                          & F1  & \res{70.12}{0.59} & \res{67.20}{0.27} & \res{55.05}{0.54} & \res{14.30}{0.26} & \res{16.50}{0.40} & \res{16.94}{0.73} & \res{27.89}{0.94} & \res{7.65}{0.11} & \res{1.65}{0.01} & \res{14.14}{0.12} & \res{5.71}{0.10} & \res{12.58}{0.21} \\
\cmidrule{2-15}
& \multirow{2}{*}{MinCut}   & ARI & \res{28.60}{2.86} & \res{50.99}{2.33} & \res{39.89}{2.65} & \res{1.22}{0.30} & \res{3.73}{0.66} & \res{17.47}{0.98} & \res{25.42}{1.67} & \res{5.56}{0.15} & \res{0.61}{0.04} & \res{10.64}{0.32} & \res{2.00}{0.20} & -- \\
&                          & F1  & \res{51.88}{6.97} & \res{66.99}{2.97} & \res{52.57}{2.83} & \res{9.54}{0.68} & \res{21.22}{1.17} & \res{21.85}{0.73} & \res{26.39}{5.74} & \res{8.04}{0.12} & \res{2.30}{0.05} & \res{17.53}{0.54} & \res{6.92}{0.46} & -- \\
\cmidrule{2-15}
& \multirow{2}{*}{DMoN}     & ARI & \res{33.35}{3.35} & \res{51.55}{2.41} & \res{39.81}{0.21} & \res{2.05}{0.28} & \res{5.01}{0.52} & \res{15.44}{0.37} & \res{32.39}{1.40} & \res{4.93}{0.02} & \second{1.17}{0.04} & \res{10.93}{0.41} & \res{2.43}{0.21} & -- \\
&                          & F1  & \res{49.57}{3.84} & \res{70.86}{2.44} & \res{54.98}{0.37} & \res{10.58}{0.44} & \second{22.76}{0.48} & \res{20.37}{0.61} & \res{26.52}{0.76} & \res{8.97}{0.14} & \best{3.04}{0.04} & \res{16.43}{1.13} & \res{7.80}{0.21} & -- \\
\cmidrule{2-15}
& \multirow{2}{*}{Neuromap} & ARI & \res{36.69}{3.25} & \res{53.01}{3.67} & \res{43.52}{3.94} & \res{1.45}{0.07} & \res{4.99}{1.28} & \res{21.92}{1.08} & \res{25.32}{0.26} & \best{9.85}{0.36} & \best{1.84}{0.11} & \res{16.52}{1.56} & \second{3.31}{0.56} & -- \\
&                          & F1  & \res{54.69}{3.95} & \res{69.13}{3.76} & \res{40.41}{0.71} & \res{9.37}{0.18} & \res{22.39}{1.57} & \res{21.29}{0.84} & \res{25.70}{0.43} & \res{6.20}{0.19} & \res{2.62}{0.14} & \res{13.09}{1.04} & \res{7.96}{0.39} & -- \\

\bottomrule
\end{tabular}
\end{adjustbox}
\end{table*}

\begin{table*}[t]
\centering
\caption{
Clustering performance comparison measured by Modularity (Mod) $\uparrow$ and Conductance (Cond) $\downarrow$ ($\%$) (Mean $\pm$ SD). The \colorbox{colorBest}{\textbf{best}} and \colorbox{colorSecond}{second-best} results are highlighted. "--" denotes OOM errors as these methods strictly require full-graph processing.}
\label{tab:mod_cond}
\begin{adjustbox}{width=\textwidth, center}
\begin{tabular}{lll cccccccccc cc}
\toprule
~ & \multirow{2}{*}{\textbf{Model}} & \multirow{2}{*}{\textbf{Metric}} & \multicolumn{2}{c}{\textbf{Tiny}} & \multicolumn{3}{c}{\textbf{Small}} & \multicolumn{3}{c}{\textbf{Medium}} & \multicolumn{3}{c}{\textbf{Large}} & \multicolumn{1}{c}{\textbf{Massive}} \\
\cmidrule(lr){4-5} \cmidrule(lr){6-8} \cmidrule(lr){9-11} \cmidrule(lr){12-14} \cmidrule(lr){15-15}
& & & \textbf{Cora} & \textbf{Photo} & \textbf{Physics} & \textbf{HM} & \textbf{Flickr} & \textbf{ArXiv} & \textbf{Reddit} & \textbf{MAG} & \textbf{Pokec} & \textbf{Products} & \textbf{WebTop.} & \textbf{Papers.} \\
\midrule

\multirow{4}{*}{\rotatebox{90}{Traditional}} 
& \multirow{2}{*}{KMeans}   & Mod & \res{18.73}{3.05} & \res{16.85}{0.40} & \res{51.56}{0.07} & \res{2.84}{0.05} & \res{0.90}{0.08} & \res{13.23}{0.31} & \res{4.48}{0.15} & \res{14.53}{0.06} & \res{0.89}{0.04} & \res{28.88}{0.28} & \res{0.56}{0.19} & \res{15.79}{0.09} \\
&                          & Cond & \res{57.49}{5.83} & \res{66.60}{0.47} & \res{24.55}{0.04} & \res{90.80}{0.42} & \res{74.78}{0.69} & \res{83.62}{0.38} & \res{91.60}{0.32} & \res{85.02}{0.07} & \res{98.05}{0.05} & \res{67.92}{0.40} & \res{90.76}{0.35} & \res{83.18}{0.10} \\
\cmidrule{2-15}
& \multirow{2}{*}{Node2Vec} & Mod & \res{71.52}{0.15} & \best{71.65}{0.14} & \res{59.19}{0.00} & \res{-5.89}{0.08} & \res{41.93}{0.01} & \res{55.48}{0.56} & \res{61.41}{0.58} & \res{45.08}{0.26} & \res{30.44}{0.16} & \res{77.17}{0.66} & \res{3.24}{0.12} & \res{42.23}{0.48} \\
&                          & Cond & \res{11.99}{0.14} & \res{10.76}{0.10} & \res{15.54}{0.00} & \res{81.80}{0.25} & \second{40.24}{0.01} & \res{39.80}{0.65} & \res{35.05}{0.59} & \res{54.42}{0.28} & \res{68.84}{0.16} & \res{20.23}{0.66} & \res{69.74}{0.17} & \res{55.66}{0.54} \\

\midrule
\multirow{6}{*}{\rotatebox{90}{Non-Parametric}} 
& \multirow{2}{*}{SSGC}    & Mod & \res{72.60}{0.35} & \res{70.13}{1.91} & \res{57.59}{4.45} & \res{4.85}{0.03} & \res{20.86}{0.30} & \res{60.21}{0.12} & \res{33.83}{1.19} & \res{64.64}{0.07} & \res{16.94}{0.41} & \res{80.58}{0.16} & \res{17.82}{0.79} & -- \\
&                          & Cond & \res{10.84}{0.09} & \best{9.04}{1.23} & \res{11.44}{2.72} & \res{69.42}{0.33} & \res{60.13}{0.44} & \second{26.94}{0.15} & \res{61.75}{1.39} & \res{34.61}{0.05} & \res{80.09}{0.54} & \res{16.41}{0.11} & \res{62.67}{1.82} & -- \\
\cmidrule{2-15}
& \multirow{2}{*}{SAGSC}   & Mod & \res{55.76}{0.05} & \res{65.11}{0.01} & \res{45.40}{0.01} & \res{-2.96}{0.39} & \res{24.73}{0.03} & \res{53.04}{2.24} & \second{68.43}{0.63} & \second{70.88}{0.31} & \best{50.01}{0.27} & \best{84.56}{0.17} & \res{16.81}{1.17} & -- \\
&                          & Cond & \res{27.61}{0.06} & \res{17.16}{0.01} & \res{20.60}{0.01} & \res{75.53}{0.60} & \res{57.19}{0.03} & \res{38.27}{2.81} & \res{27.46}{0.64} & \second{28.43}{0.34} & \best{49.02}{0.28} & \best{12.62}{0.19} & \res{67.47}{0.71} & -- \\
\cmidrule{2-15}
& \multirow{2}{*}{MS2CAG}  & Mod & \second{74.67}{0.15} & \res{71.08}{0.05} & \res{49.60}{0.02} & \res{2.50}{0.10} & \second{44.40}{0.03} & \second{64.06}{0.27} & \res{48.58}{1.89} & \res{63.42}{0.41} & \res{3.54}{0.17} & \second{83.80}{0.14} & \best{35.33}{0.46} & -- \\
&                          & Cond & \res{9.49}{0.09} & \res{10.11}{0.08} & \second{6.27}{0.02} & \res{90.94}{0.22} & \best{39.11}{0.03} & \best{26.87}{0.33} & \res{47.72}{2.04} & \res{35.98}{0.42} & \res{95.76}{0.19} & \second{13.58}{0.13} & \res{54.68}{0.69} & -- \\

\midrule
\multirow{12}{*}{\rotatebox{90}{Deep Decoupled}} 
& \multirow{2}{*}{GAE}     & Mod & \res{72.39}{0.11} & \res{66.94}{0.17} & \res{49.97}{0.02} & \res{3.49}{0.18} & \res{21.66}{0.53} & \res{58.22}{0.26} & \res{36.96}{0.35} & \res{54.26}{0.22} & \res{11.66}{0.07} & \res{62.17}{0.17} & \res{5.24}{0.24} & \res{25.79}{0.05} \\
&                          & Cond & \res{11.72}{0.11} & \res{15.27}{0.27} & \res{8.23}{0.01} & \res{88.66}{0.22} & \res{47.06}{2.26} & \res{37.82}{0.39} & \res{58.48}{0.32} & \res{45.07}{0.23} & \res{85.78}{0.08} & \res{35.36}{0.16} & \res{80.57}{0.21} & \res{72.38}{0.05} \\
\cmidrule{2-15}
& \multirow{2}{*}{DGI}     & Mod & \res{71.49}{0.28} & \res{64.99}{0.37} & \res{48.44}{0.09} & \res{5.53}{0.10} & \res{6.81}{0.25} & \res{54.30}{0.50} & \res{62.19}{0.27} & \res{56.43}{0.48} & \res{6.65}{0.04} & \res{61.54}{0.25} & \res{-1.07}{0.44} & \res{32.21}{0.20} \\
&                          & Cond & \res{12.43}{0.78} & \res{18.24}{1.07} & \res{6.80}{0.11} & \res{67.22}{1.10} & \res{75.68}{0.25} & \res{39.15}{0.69} & \res{33.44}{0.39} & \res{42.76}{0.50} & \res{92.40}{0.03} & \res{35.60}{0.23} & \res{88.34}{0.23} & \res{66.61}{0.19} \\
\cmidrule{2-15}
& \multirow{2}{*}{CCASSG}  & Mod & \res{73.46}{0.42} & \res{68.54}{1.69} & \res{49.12}{0.03} & \res{6.48}{0.76} & \res{-4.17}{0.46} & \res{49.95}{0.16} & \res{37.92}{0.34} & \res{57.57}{0.16} & \res{0.84}{0.00} & \res{74.64}{0.17} & \res{0.76}{0.71} & \best{50.06}{0.28} \\
&                          & Cond & \res{10.18}{0.22} & \res{13.24}{2.12} & \res{6.91}{0.02} & \res{68.21}{1.45} & \res{85.64}{0.10} & \res{45.83}{0.24} & \res{57.98}{0.39} & \res{41.65}{0.17} & \res{98.34}{0.01} & \res{22.33}{0.20} & \res{88.46}{0.93} & \best{46.69}{0.34} \\
\cmidrule{2-15}
& \multirow{2}{*}{S3GC}    & Mod & \res{74.49}{0.07} & \res{71.04}{0.07} & \res{49.72}{0.01} & \second{11.14}{0.30} & \res{39.96}{0.01} & \res{57.98}{0.39} & \best{70.94}{1.25} & \res{49.45}{0.26} & \res{13.53}{0.15} & \res{73.93}{0.33} & \res{29.06}{0.26} & \res{37.52}{0.21} \\
&                          & Cond & \second{9.12}{0.20} & \res{9.44}{0.22} & \res{6.64}{0.01} & \res{64.03}{0.90} & \res{45.30}{0.01} & \res{35.26}{0.61} & \second{24.45}{1.41} & \res{49.37}{0.29} & \res{85.14}{0.16} & \res{23.28}{0.42} & \res{60.56}{0.45} & \res{60.15}{0.25} \\
\cmidrule{2-15}
& \multirow{2}{*}{NS4GC}   & Mod & \best{75.08}{0.25} & \second{71.40}{0.01} & \res{48.94}{0.01} & \res{5.11}{0.19} & \res{-5.15}{0.05} & \res{62.37}{0.27} & \res{41.98}{0.14} & \res{62.80}{0.21} & \res{15.98}{0.09} & \res{77.89}{0.18} & \res{24.51}{0.06} & \res{42.44}{0.55} \\
&                          & Cond & \best{8.49}{0.12} & \second{9.12}{0.04} & \best{5.61}{0.01} & \res{85.63}{0.38} & \res{74.84}{0.06} & \res{29.25}{0.67} & \res{54.28}{0.15} & \res{36.40}{0.22} & \res{82.72}{0.10} & \res{19.34}{0.20} & \res{52.27}{0.69} & \res{54.51}{0.66} \\
\cmidrule{2-15}
& \multirow{2}{*}{MAGI}    & Mod & \res{71.19}{0.06} & \res{68.69}{0.09} & \res{60.53}{0.19} & \res{-4.15}{0.18} & \best{47.35}{0.08} & \res{61.71}{0.23} & \res{55.80}{0.43} & \res{64.73}{0.18} & \res{19.64}{0.07} & \res{68.51}{0.03} & \second{34.96}{1.23} & \second{47.63}{0.61} \\
&                          & Cond & \res{12.31}{0.05} & \res{13.74}{0.11} & \res{13.63}{0.12} & \res{82.99}{0.46} & \best{36.42}{0.37} & \res{30.04}{0.34} & \res{39.97}{0.56} & \res{34.55}{0.19} & \res{79.44}{0.08} & \res{28.80}{0.03} & \res{48.18}{0.71} & \second{49.31}{0.78} \\

\midrule
\multirow{10}{*}{\rotatebox{90}{Deep Joint}} 
& \multirow{2}{*}{DAEGC}   & Mod & \res{73.45}{0.62} & \res{65.34}{0.01} & \second{61.95}{0.02} & \res{4.40}{0.21} & \res{19.45}{0.09} & \res{62.65}{0.18} & \res{49.83}{1.08} & \res{61.72}{1.31} & \res{10.26}{0.31} & \res{35.59}{3.73} & \res{27.16}{1.13} & \res{41.13}{0.57} \\
&                          & Cond & \res{11.08}{0.65} & \res{16.79}{0.02} & \res{14.35}{0.01} & \res{79.05}{0.70} & \res{64.14}{0.21} & \res{33.13}{0.24} & \best{21.36}{1.87} & \res{37.60}{1.31} & \res{87.47}{0.29} & \res{58.71}{3.05} & \second{46.11}{2.53} & \res{52.17}{1.94} \\
\cmidrule{2-15}
& \multirow{2}{*}{DinkNet} & Mod & \res{72.26}{0.16} & \res{64.98}{0.21} & \res{60.92}{0.06} & \res{5.93}{0.22} & \res{4.37}{0.52} & \res{31.87}{0.36} & \res{42.73}{0.38} & \res{34.56}{0.27} & \res{12.47}{0.34} & \res{59.75}{0.25} & \res{-2.34}{0.28} & \res{32.51}{0.43} \\
&                          & Cond & \res{11.47}{0.10} & \res{18.15}{0.22} & \res{15.08}{0.06} & \res{67.05}{0.71} & \res{69.02}{0.98} & \res{62.35}{0.47} & \res{52.26}{0.47} & \res{64.64}{0.28} & \res{84.40}{0.70} & \res{36.49}{0.34} & \res{92.27}{0.05} & \res{65.10}{0.48} \\
\cmidrule{2-15}
& \multirow{2}{*}{MinCut}  & Mod & \res{72.99}{1.28} & \res{67.00}{1.75} & \res{61.47}{0.20} & \res{4.25}{2.35} & \res{29.50}{4.82} & \res{61.80}{1.08} & \res{42.38}{1.55} & \res{68.97}{0.25} & \res{14.55}{0.43} & \res{73.38}{0.30} & \res{24.86}{1.25} & -- \\
&                          & Cond & \res{12.28}{1.23} & \res{15.52}{2.57} & \res{14.18}{0.41} & \second{26.38}{14.89} & \res{49.40}{1.61} & \res{32.05}{1.30} & \res{51.94}{5.08} & \res{30.22}{0.28} & \res{84.46}{0.44} & \res{24.17}{0.30} & \res{50.80}{2.93} & -- \\
\cmidrule{2-15}
& \multirow{2}{*}{DMoN}    & Mod & \res{71.81}{1.40} & \res{70.32}{0.28} & \best{62.90}{0.11} & \best{12.54}{0.25} & \res{37.17}{0.74} & \res{61.14}{1.19} & \res{47.78}{1.70} & \res{69.58}{0.15} & \res{22.67}{0.28} & \res{68.35}{0.53} & \res{33.90}{0.23} & -- \\
&                          & Cond & \res{12.63}{0.86} & \res{13.07}{0.20} & \res{13.91}{0.12} & \res{66.99}{0.99} & \res{46.19}{0.82} & \res{33.53}{0.88} & \res{46.65}{2.04} & \res{29.77}{0.14} & \res{76.06}{0.30} & \res{28.76}{0.43} & \res{52.97}{0.84} & -- \\
\cmidrule{2-15}
& \multirow{2}{*}{Neuromap} & Mod & \res{74.02}{0.95} & \res{69.10}{1.96} & \res{57.35}{0.33} & \res{6.29}{0.17} & \res{34.35}{2.98} & \best{64.54}{0.66} & \res{41.11}{0.27} & \best{76.25}{0.15} & \second{31.71}{0.56} & \res{70.59}{0.43} & \res{31.96}{1.39} & -- \\
&                          & Cond & \res{10.96}{0.93} & \res{12.21}{0.98} & \res{11.55}{1.62} & \best{20.28}{0.45} & \res{47.55}{2.23} & \res{27.76}{1.09} & \res{54.20}{0.32} & \best{21.75}{0.15} & \second{65.39}{0.65} & \res{23.49}{0.52} & \best{38.20}{0.72} & -- \\

\bottomrule
\end{tabular}
\end{adjustbox}
\vspace{-0.3cm}
\end{table*}

\subsection{Generalization Gap: Academia vs. Industry}
Tables~\ref{tab:nmi_acc} and~\ref{tab:ari_f1} present the clustering performance measured by supervised alignment metrics (NMI, ACC, ARI, F1). A comparative analysis reveals a stark dichotomy between performance on traditional academic datasets and industrial-scale graphs.

\vpara{The Cora Comfort Zone.} 
On classic small-scale, high-homophily datasets (\texttt{Cora}, \texttt{Photo}, \texttt{Physics}), most methods achieve high performance. 
For instance, on \texttt{Photo}, simple baselines like Node2Vec and sophisticated methods like NS4GC both achieve NMI scores $>65\%$. 
This confirms that in clean, homophilous environments, structure and attributes are highly correlated, making the clustering task relatively trivial.

\vpara{Collapse on Industrial Graphs.} 
Performance degrades largely when moving to industrial environments (\texttt{HM}, \texttt{Pokec}, \texttt{WebTopic}). 
On \texttt{Pokec} (a large social network), the best performing method (SAGSC) only achieves an NMI of $38.33\%$, while most deep learning methods (e.g., DAEGC, DGI) fail to surpass $5\%$ NMI.
This collapse is attributed to two factors:
\begin{itemize}[leftmargin=*]
    \item \textbf{Heterophily and Noise:} Industrial graphs like \texttt{HM} ($\mathcal{H}_e = 0.16$) are highly heterophilous. Methods relying on GNN encoders tend to over-smooth features across class boundaries, rendering the learned embeddings indistinguishable.
    \item \textbf{Tabular Complexity:} Unlike the clean text features, the tabular features contain heterogeneous distributions and noise. Traditional KMeans and spectral filtering methods struggle to capture the complex dependencies in this feature space without learnable non-linear transformations.
\end{itemize}

\vpara{Robustness of Deep Decoupled Methods.} 
Among the evaluated categories, Deep Decoupled methods (e.g., NS4GC, MAGI, S3GC)  demonstrate the most consistent robustness. 
By separating representation learning from clustering, these methods benefit from stable self-supervised pre-training, avoiding the \textit{clustering collapse} phenomenon often observed in joint training (where the clusterer forces the encoder into a trivial solution).

\subsection{The Metric Paradox: Labels vs. Structure}
A core premise of this benchmark is that supervised metrics are insufficient for evaluating unsupervised graph clustering. Table~\ref{tab:mod_cond} (Modularity and Conductance) highlights the disconnect between structural quality and label alignment.

\vpara{Structure vs. Semantics Misalignment.} 
We observe instances where supervised metrics and structural metrics diverge. 
For example, on \texttt{Products}, SAGSC achieves the highest Modularity (84.56\%) and lowest Conductance (12.62\%), indicating it finds extremely dense, well-separated communities. 
However, its NMI (51.78\%) is lower than NS4GC (54.63\%). 
This highlights a critical nuance: \textit{ground-truth labels do not always reflect topological communities}. 
In practical applications like fraud detection, a partition with high modularity (isolating dense cliques) may be more valuable than one aligning with broad semantic labels.

\vpara{Failure Modes in Heterophily.}
On the highly heterophilous \texttt{HM} dataset ($\mathcal{H}_e = 0.16$), standard GNN methods yield low or even negative Modularity scores, implying they fail to detect non-trivial community structures. 
Interestingly, DMoN, which directly optimizes a spectral modularity objective, achieves the highest structural quality ($\mathcal{Q}=12.54\%$), despite having mediocre supervised scores. 
This underscores the necessity of including structural metrics in evaluation; relying solely on Accuracy would mask DMoN's utility in detecting latent structure in label-poor environments.

\begin{table*}[t]
\centering
\caption{Efficiency profiling results for smaller datasets. Mem: Peak GPU Memory, Time: Total Training and Clustering Time.}
\label{tab:efficiency_1}
\small
\resizebox{\textwidth}{!}{
\begin{tabular}{ll|cc|cc|cc|cc|cc|cc}
\toprule
\multirow{2}{*}{\textbf{Category}} & \multirow{2}{*}{\textbf{Method}} & \multicolumn{2}{c|}{\textbf{Cora}} & \multicolumn{2}{c|}{\textbf{Photo}} & \multicolumn{2}{c|}{\textbf{Physics}} & \multicolumn{2}{c|}{\textbf{HM}} & \multicolumn{2}{c|}{\textbf{Flickr}} & \multicolumn{2}{c}{\textbf{ArXiv}} \\
\cmidrule(lr){3-4} \cmidrule(lr){5-6} \cmidrule(lr){7-8} \cmidrule(lr){9-10} \cmidrule(lr){11-12} \cmidrule(lr){13-14}
  & & Mem(MB) & Time(s) & Mem(GB) & Time(s) & Mem(GB) & Time(m) & Mem(GB) & Time(m) & Mem(GB) & Time(m) & Mem(GB) & Time(m) \\
\midrule
\multirow{2}{*}{Traditional} 
  & KMeans & - & 0.28 & - & 0.69 & - & 0.19 & - & 0.05 & - & 0.08 & - & 0.32 \\
  & Node2Vec & 46.0 & 24.57 & 0.07 & 27.96 & 0.16 & 15.52 & 0.19 & 1.50 & 0.36 & 20.97 & 0.49 & 5.05 \\
\midrule
\multirow{3}{*}{\shortstack[l]{Non-\\Parametric}} 
  & SSGC & - & 0.56 & - & 0.62 & - & 2.25 & - & 0.04 & - & 0.10 & - & 0.17 \\
  & SAGSC & - & 0.86 & - & 1.02 & - & 2.13 & - & 0.07 & - & 0.04 & - & 0.28 \\
  & MS2CAG & - & 0.78 & - & 0.85 & - & 0.15 & - & 0.03 & - & 0.03 & - & 0.06 \\
\midrule
\multirow{6}{*}{\shortstack[l]{Deep\\Decoupled}} 
  & GAE & 60.0 & 32.03 & 1.48 & 178.93 & 2.86 & 1.17 & 0.38 & 0.81 & 2.86 & 2.56 & 1.76 & 12.77 \\
  & DGI & 152.0 & 15.50 & 1.51 & 11.14 & 5.40 & 1.55 & 16.55 & 0.54 & 6.25 & 1.12 & 15.11 & 2.76 \\
  & CCASSG & 46.0 & 0.46 & 1.17 & 27.92 & 4.36 & 0.08 & 26.30 & 0.18 & 1.51 & 0.09 & 13.55 & 3.04 \\
  & S3GC & 438.0 & 124.74 & 1.73 & 148.44 & 11.59 & 50.11 & 0.35 & 0.98 & 2.12 & 15.49 & 1.21 & 3.91 \\
  & NS4GC & 278.0 & 9.77 & 1.46 & 64.36 & 9.96 & 3.12 & 6.53 & 4.47 & 7.10 & 0.85 & 11.69 & 22.94 \\
  & MAGI & 210.0 & 51.25 & 1.06 & 18.68 & 12.66 & 26.22 & 1.24 & 41.84 & 14.97 & 43.31 & 30.54 & 28.80 \\
\midrule
\multirow{5}{*}{\shortstack[l]{Deep\\Joint}} 
  & DAEGC & 68.0 & 35.80 & 2.17 & 156.84 & 6.13 & 5.56 & 22.29 & 1.10 & 4.24 & 4.87 & 6.13 & 9.06 \\
  & DinkNet & 192.0 & 13.31 & 2.95 & 36.07 & 5.40 & 1.93 & 16.56 & 0.14 & 6.45 & 1.40 & 19.88 & 3.72 \\
  & MinCut & 62.0 & 5.51 & 0.25 & 6.78 & 1.34 & 0.23 & 16.55 & 5.11 & 0.99 & 0.34 & 3.95 & 1.00 \\
  & DMoN & 62.0 & 3.58 & 0.25 & 6.99 & 1.34 & 0.21 & 16.55 & 5.08 & 0.99 & 0.34 & 3.95 & 1.37 \\
  & Neuromap & 62.0 & 4.73 & 0.80 & 12.89 & 1.34 & 0.23 & 16.55 & 5.17 & 1.74 & 0.47 & 3.96 & 1.62 \\
\bottomrule
\end{tabular}}
\end{table*}

\begin{table*}[t]
\centering
\caption{Efficiency profiling results for larger datasets. Mem: Peak GPU Memory, Time: Total Training and Clustering Time.}
\label{tab:efficiency_2}
\small
\resizebox{\textwidth}{!}{
\begin{tabular}{ll|cc|cc|cc|cc|cc|cc}
\toprule
\multirow{2}{*}{\textbf{Category}} & \multirow{2}{*}{\textbf{Method}} & \multicolumn{2}{c|}{\textbf{Reddit}} & \multicolumn{2}{c|}{\textbf{MAG}} & \multicolumn{2}{c|}{\textbf{Pokec}} & \multicolumn{2}{c|}{\textbf{Products}} & \multicolumn{2}{c|}{\textbf{WebTopic}} & \multicolumn{2}{c}{\textbf{Papers100M}} \\
\cmidrule(lr){3-4} \cmidrule(lr){5-6} \cmidrule(lr){7-8} \cmidrule(lr){9-10} \cmidrule(lr){11-12} \cmidrule(lr){13-14}
 & & Mem(GB) & Time(m) & Mem(GB) & Time(m) & Mem(GB) & Time(m) & Mem(GB) & Time(m) & Mem(GB) & Time(m) & Mem(GB) & Time(h) \\
\midrule
\multirow{2}{*}{Traditional} & KMeans & - & 0.27 & - & 6.59 & - & 3.09 & - & 5.87 & - & 3.98 & - & 0.05 \\
 & Node2Vec & 0.70 & 22.67 & 1.90 & 77.58 & 4.04 & 236.53 & 15.05 & 64.94 & 11.80 & 75.92 & 7.20 & 12.38 \\
\midrule
\multirow{3}{*}{\shortstack[l]{Non-\\Parametric}} & SSGC & - & 5.36 & - & 6.51 & - & 4.43 & - & 8.05 & - & 10.38 & - & - \\
 & SAGSC & - & 5.99 & - & 23.35 & - & 4.72 & - & 5.29 & - & 2.77 & - & - \\
 & MS2CAG & - & 0.16 & - & 15.10 & - & 13.68 & - & 1.38 & - & 16.81 & - & - \\
\midrule
\multirow{5}{*}{\shortstack[l]{Deep\\Decoupled}} & GAE & 14.18 & 30.58 & 10.16 & 26.54 & 27.97 & 10.16 & 13.33 & 26.12 & 1.63 & 6.18 & 7.58 & 0.67 \\
 & DGI & 24.16 & 1.93 & 22.46 & 10.54 & 30.90 & 10.37 & 29.52 & 50.32 & 8.96 & 6.16 & 30.89 & 2.42 \\
 & CCASSG & 27.50 & 1.21 & 28.64 & 31.14 & 30.75 & 117.43 & 30.30 & 45.16 & 6.16 & 122.30 & 31.09 & 1.29 \\
 & S3GC & 2.82 & 32.38 & 5.12 & 41.78 & 9.01 & 71.17 & 29.46 & 29.68 & 18.98 & 145.98 & 15.85 & 5.85 \\
 & NS4GC & 10.90 & 12.84 & 17.71 & 41.04 & 30.12 & 55.88 & 24.10 & 37.24 & 8.95 & 30.75 & 13.97 & 1.24 \\
 & MAGI & 20.41 & 56.98 & 30.77 & 86.99 & 21.44 & 176.43 & 31.08 & 375.99 & 5.23 & 115.36 & 29.28 & 6.04 \\
\midrule
\multirow{5}{*}{\shortstack[l]{Deep\\Joint}} & DAEGC & 31.20 & 14.28 & 26.18 & 27.52 & 30.90 & 14.28 & 15.37 & 37.28 & 11.71 & 11.11 & 15.96 & 1.40 \\
 & DinkNet & 24.16 & 4.49 & 23.66 & 5.46 & 24.84 & 6.18 & 30.63 & 85.27 & 29.98 & 4.27 & 31.11 & 5.13 \\
 & MinCut & 18.56 & 6.73 & 18.16 & 16.01 & 19.32 & 24.15 & 30.14 & 32.55 & 28.27 & 8.37 & - & - \\
 & DMoN & 18.56 & 1.22 & 18.16 & 15.46 & 19.32 & 22.91 & 30.14 & 22.88 & 28.27 & 6.75 & - & - \\
 & Neuromap & 18.57 & 6.73 & 18.16 & 14.56 & 19.32 & 23.07 & 24.50 & 23.00 & 28.27 & 4.97 & - & - \\
\bottomrule
\end{tabular}}
\vspace{-0.3cm}
\end{table*}

\subsection{Scalability and Efficiency Profiling}
Finally, we analyze the computational efficiency of these methods, enabled by the \texttt{PyAGC} mini-batch framework. 
Tables~\ref{tab:efficiency_1} and \ref{tab:efficiency_2} detail peak memory usage and total runtime.

\vpara{Breaking the Memory Wall.}
Standard implementations of methods like DAEGC or NS4GC typically scale quadratically or require the full adjacency matrix, making them infeasible for graphs larger than \texttt{Physics}. 
Our mini-batch implementations clamp GPU memory usage effectively. 
As shown in Table~\ref{tab:efficiency_2}, training NS4GC on \texttt{Papers100M} consumes only $\approx$14GB of GPU memory, well within the limits of a 32GB V100, whereas the full-batch equivalent would require terabytes of RAM.

\vpara{Training Throughput.}
\texttt{PyAGC} achieves production-grade throughput. 
We successfully train deep clustering models on \texttt{Papers100M} (111M nodes) in remarkably short windows: GAE completes an epoch in 0.67 hours, and contrastive methods like NS4GC take 1.24 hours. 
This proves that deep graph clustering is no longer limited to toy datasets and is ready for high-frequency industrial retraining cycles.
Among deep joint methods, DMoN and Neuromap are particularly efficient due to their compact matrix formulations, processing \texttt{WebTopic} in under 7 minutes.

\vpara{Trade-offs.}
Non-parametric methods like MS2CAG are extremely fast on medium datasets (0.16 min on \texttt{Reddit}) but hit a hard scalability wall on Massive graphs. 
Conversely, deep mini-batch methods incur a time overhead due to sampling but offer linear scalability to virtually infinite graph sizes. 
This distinction allows practitioners to choose the right tool: non-parametric solvers for rapid prototyping on medium graphs, and mini-batch deep models for massive-scale deployment. 
Additionally, methods with fast convergence and strong performance (e.g., CCASSG: 1.29h, NMI=53.82\% on \texttt{Papers100M}) are inherently preferable in high-frequency industrial retraining cycles.

\vpara{\emph{Supplementary Analysis.}}
Appendix~\ref{app:datasets} details the 12 benchmark datasets, Appendix~\ref{app:baselines} describes the evaluated algorithms, and Appendix~\ref{app:implementation_details} reports implementation and training details. 
We further include ablation studies on graph encoders and neural tabular encoding in Appendices~\ref{app:encoder_ablation} and~\ref{app:tabular_encoder}, respectively, followed by limitations, ethical considerations, and broader impacts in Appendix~\ref{app:misc}.

\section{Conclusion and Future Work}
\label{sec:conclusion}
\vspace{-0.2cm}
In this work, we bridged the widening chasm between academic research and industrial application in AGC. 
We introduced \texttt{PyAGC}, a benchmark library that systematizes the field under the Encode-Cluster-Optimize framework and democratizes access to massive-scale evaluations.
Our rigorous benchmarking across 12 datasets and 17 representative algorithms yields three critical insights.
By releasing this benchmark, we invite researchers to test their methods against the realities of scale and complexity, ultimately advancing AGC towards reliable, high-stakes deployment.

Looking forward, \texttt{PyAGC} opens several avenues for future research. 
The community must develop robust encoders that can handle heterophily and tabular noise without over-smoothing~\cite{FT-Transformer, TFM4GAD}.
As industrial applications are often label-scarce, reliable unsupervised model selection remains an urgent open problem~\cite{AGCSurvey}.
Beyond the static homogeneous graphs in this benchmark, clustering should extend to heterogeneous, multiplex, and temporal graphs~\cite{HG2M, MGFNN, BenchTGC}, including hierarchical clustering that recovers community structure at multiple resolutions rather than a single flat partition~\cite{TIER}.
A complementary direction is clustering-oriented graph foundation models: pretrained encoders whose objectives and adaptation protocols target partition quality, so that one model can transfer across graphs instead of training a separate AGC method on each dataset~\cite{OFAG}.

\begin{ack}
This work was supported by the National Science and Technology Major Project (2026ZD16011200), National Key Research and Development Plan of China (2023YFB4502305), and the Ant Group Research Intern Program.
\end{ack}

\bibliographystyle{plainnat}
\bibliography{main}


\appendix

\section*{Appendix Table of Contents}
\label{sec:appendix_toc}

To facilitate navigation, we organize the supplementary materials as follows:

\begin{itemize}[leftmargin=1.5em, itemsep=0.2em]
    \item \textbf{Appendix~\ref{app:datasets}: Dataset Descriptions.} Detailed descriptions of the 12 benchmark datasets across domains, scales, feature modalities, and homophily levels.
    \item \textbf{Appendix~\ref{app:baselines}: Evaluated Algorithms.} Detailed descriptions of the 17 representative AGC methods under the Encode--Cluster--Optimize taxonomy.
    \item \textbf{Appendix~\ref{app:implementation_details}: Implementation Details.} Hardware environment, reproducibility settings, hyperparameter management, and training protocol.
    \item \textbf{Appendix~\ref{app:encoder_ablation}: Graph Encoder Ablation.} Analysis of how different GNN encoder backbones affect clustering performance.
    \item \textbf{Appendix~\ref{app:tabular_encoder}: Neural Tabular Encoding Ablation.} Study of learnable tabular feature encoding for industrial graphs.
    \item \textbf{Appendix~\ref{app:misc}: Miscellaneous.} Limitations, ethical considerations, and broader impacts.
\end{itemize}

\clearpage

\begin{table*}[t]
\centering
\caption{
Clustering performance comparison measured by Homogeneity (Homo) and Completeness (Comp) ($\%$) (Mean $\pm$ SD). The \colorbox{colorBest}{\textbf{best}} and \colorbox{colorSecond}{second-best} results are highlighted. "--" denotes OOM errors as these methods strictly require full-graph processing.}
\label{tab:benchmark_results}
\begin{adjustbox}{width=\textwidth, center}
\begin{tabular}{llll cccccccccc cc}
\toprule
~ & \multirow{2}{*}{\textbf{Model}} & \multirow{2}{*}{\textbf{Metric}} & \multicolumn{2}{c}{\textbf{Tiny}} & \multicolumn{3}{c}{\textbf{Small}} & \multicolumn{3}{c}{\textbf{Medium}} & \multicolumn{3}{c}{\textbf{Large}} & \multicolumn{1}{c}{\textbf{Massive}} \\
\cmidrule(lr){4-5} \cmidrule(lr){6-8} \cmidrule(lr){9-11} \cmidrule(lr){12-14} \cmidrule(lr){15-15}
& & & \textbf{Cora} & \textbf{Photo} & \textbf{Physics} & \textbf{HM} & \textbf{Flickr} & \textbf{ArXiv} & \textbf{Reddit} & \textbf{MAG} & \textbf{Pokec} & \textbf{Products} & \textbf{WebTop} & \textbf{Papers.} \\
\midrule

\multirow{4}{*}{\rotatebox{90}{Trad.}} 
& \multirow{2}{*}{KMeans}   & Homo & \res{12.84}{3.78} & \res{32.66}{0.61} & \res{55.66}{0.17} & \res{10.80}{0.13} & \res{1.20}{0.07} & \res{25.08}{0.02} & \res{11.49}{0.16} & \res{30.56}{0.05} & \res{1.44}{0.00} & \res{33.93}{0.19} & \res{2.83}{0.06} & \res{41.09}{0.07} \\
&                          & Comp & \res{15.19}{5.37} & \res{30.80}{0.40} & \res{48.79}{0.10} & \res{9.60}{0.10}  & \res{1.22}{0.07} & \res{20.54}{0.04} & \res{10.69}{0.13} & \res{26.38}{0.06} & \res{1.37}{0.00} & \res{25.68}{0.10} & \res{2.16}{0.05} & \res{34.95}{0.07} \\
\cmidrule{2-15}
& \multirow{2}{*}{Node2Vec} & Homo & \res{44.76}{1.49} & \res{68.02}{1.20} & \res{58.81}{0.01} & \res{7.97}{0.13}  & \res{6.04}{0.00} & \res{42.93}{0.16} & \second{81.69}{0.35} & \res{40.74}{0.03} & \second{33.71}{0.02} & \res{60.22}{0.18} & \res{6.73}{0.10} & \res{55.43}{0.06} \\
&                          & Comp & \res{45.14}{1.49} & \res{64.13}{0.91} & \res{51.55}{0.01} & \res{7.26}{0.13}  & \res{5.20}{0.00} & \res{35.61}{0.15} & \second{76.93}{0.44} & \res{35.13}{0.03} & \second{30.32}{0.02} & \res{44.18}{0.24} & \res{5.09}{0.08} & \res{47.73}{0.05} \\
\midrule

\multirow{6}{*}{\rotatebox{90}{Non-Para.}} 
& \multirow{2}{*}{SSGC}    & Homo & \res{52.39}{1.18} & \res{71.05}{2.42} & \res{68.58}{2.64} & \res{12.80}{0.11} & \res{4.80}{0.23} & \res{49.16}{0.12} & \res{52.41}{0.27} & \best{44.59}{0.03} & \res{4.36}{0.02} & \res{61.17}{0.28} & \res{4.49}{0.13} & -- \\
&                          & Comp & \res{51.33}{0.99} & \res{70.49}{0.35} & \res{60.92}{5.50} & \res{11.57}{0.10} & \res{4.20}{0.17} & \res{43.43}{0.19} & \res{50.83}{0.52} & \second{38.84}{0.04} & \res{4.31}{0.01} & \res{45.28}{0.13} & \res{3.47}{0.10} & -- \\
\cmidrule{2-15}
& \multirow{2}{*}{SAGSC}   & Homo & \res{45.48}{0.06} & \res{60.20}{0.02} & \res{60.09}{0.04} & \res{13.24}{0.12} & \res{7.44}{0.00} & \res{47.53}{0.24} & \res{81.80}{0.29} & \res{42.92}{0.03} & \best{40.18}{0.02} & \res{61.25}{0.29} & \res{10.52}{0.30} & -- \\
&                          & Comp & \res{43.27}{0.06} & \res{56.71}{0.01} & \res{52.35}{0.03} & \res{11.79}{0.14} & \res{6.36}{0.00} & \res{39.72}{0.18} & \res{78.31}{0.46} & \res{37.71}{0.03} & \best{36.64}{0.03} & \res{44.85}{0.22} & \second{8.04}{0.24} & -- \\
\cmidrule{2-15}
& \multirow{2}{*}{MS2CAG}  & Homo & \res{54.31}{0.67} & \best{73.73}{0.85} & \res{73.34}{0.04} & \res{10.37}{0.38} & \res{7.98}{0.02} & \res{47.77}{0.21} & \res{75.15}{0.39} & \res{43.24}{0.07} & \res{2.91}{0.05} & \res{60.17}{0.40} & \res{9.02}{0.10} & -- \\
&                          & Comp & \res{52.98}{0.62} & \second{71.28}{0.74} & \res{71.58}{0.04} & \res{9.20}{0.35} & \res{6.76}{0.02} & \res{41.12}{0.16} & \res{70.68}{0.68} & \res{37.89}{0.06} & \res{2.64}{0.04} & \res{43.83}{0.26} & \res{6.82}{0.07} & -- \\
\midrule

\multirow{12}{*}{\rotatebox{90}{Deep Decoupled}} 
& \multirow{2}{*}{GAE}     & Homo & \res{51.28}{0.07} & \res{62.97}{0.15} & \res{70.35}{0.02} & \second{14.44}{0.13} & \res{3.87}{0.04} & \res{45.47}{0.15} & \res{47.14}{0.28} & \res{42.38}{0.04} & \res{3.92}{0.05} & \res{50.81}{0.10} & \res{4.39}{0.17} & \res{46.13}{0.02} \\
&                          & Comp & \res{48.96}{0.07} & \res{59.83}{0.13} & \res{66.08}{0.04} & \second{12.83}{0.11} & \res{4.29}{0.14} & \res{37.11}{0.15} & \res{44.73}{0.26} & \res{36.64}{0.03} & \res{3.66}{0.04} & \res{36.60}{0.08} & \res{3.49}{0.12} & \res{39.99}{0.02} \\
\cmidrule{2-15}
& \multirow{2}{*}{DGI}     & Homo & \res{56.81}{0.52} & \res{68.40}{0.12} & \second{75.11}{0.34} & \res{12.20}{0.08} & \res{7.58}{0.01} & \res{46.46}{0.08} & \res{74.49}{0.15} & \res{42.33}{0.02} & \res{4.35}{0.01} & \res{48.81}{0.19} & \res{6.82}{0.03} & \res{53.59}{0.07} \\
&                          & Comp & \res{55.66}{1.27} & \res{67.16}{0.99} & \second{73.68}{0.36} & \res{11.18}{0.04} & \res{6.30}{0.02} & \res{38.81}{0.09} & \res{71.43}{0.15} & \res{36.82}{0.03} & \res{4.05}{0.01} & \res{35.76}{0.10} & \res{5.22}{0.03} & \res{45.62}{0.07} \\
\cmidrule{2-15}
& \multirow{2}{*}{CCASSG}  & Homo & \res{59.06}{1.03} & \res{65.99}{2.80} & \res{72.24}{0.04} & \res{12.49}{0.15} & \res{5.08}{0.02} & \res{49.50}{0.06} & \res{50.80}{0.10} & \res{43.33}{0.00} & \res{1.38}{0.01} & \res{59.33}{0.39} & \res{4.18}{0.19} & \best{57.35}{0.08} \\
&                          & Comp & \res{58.43}{0.76} & \res{63.15}{3.63} & \res{69.69}{0.05} & \res{11.42}{0.15} & \res{4.32}{0.01} & \res{40.73}{0.05} & \res{48.51}{0.05} & \res{37.85}{0.03} & \res{1.32}{0.01} & \res{44.55}{0.31} & \res{3.47}{0.18} & \best{50.70}{0.04} \\
\cmidrule{2-15}
& \multirow{2}{*}{S3GC}    & Homo & \res{55.72}{0.81} & \res{69.32}{2.19} & \res{72.21}{0.05} & \res{12.12}{0.08} & \best{8.72}{0.00} & \second{51.42}{0.12} & \best{83.23}{0.19} & \res{42.18}{0.02} & \res{6.27}{0.01} & \second{62.92}{0.12} & \res{9.07}{0.03} & \res{46.90}{0.07} \\
&                          & Comp & \res{55.18}{1.42} & \res{67.25}{2.26} & \res{69.58}{0.05} & \res{11.06}{0.09} & \second{7.11}{0.00} & \second{43.47}{0.12} & \best{83.67}{0.62} & \res{37.68}{0.04} & \res{5.83}{0.00} & \second{46.44}{0.19} & \res{6.89}{0.02} & \res{41.76}{0.09} \\
\cmidrule{2-15}
& \multirow{2}{*}{NS4GC}   & Homo & \best{59.88}{0.33} & \second{73.47}{0.88} & \best{75.91}{0.07} & \best{16.21}{0.17} & \res{6.78}{0.01} & \best{52.55}{0.17} & \res{58.15}{0.05} & \second{44.50}{0.02} & \res{7.50}{0.04} & \best{64.27}{0.20} & \best{11.55}{0.05} & \res{52.43}{0.05} \\
&                          & Comp & \best{58.93}{0.66} & \best{71.79}{0.71} & \best{74.85}{0.09} & \best{14.45}{0.18} & \res{5.69}{0.01} & \best{44.83}{0.31} & \res{55.34}{0.08} & \best{39.13}{0.02} & \res{6.91}{0.04} & \best{47.51}{0.11} & \best{8.89}{0.05} & \res{47.47}{0.08} \\
\cmidrule{2-15}
& \multirow{2}{*}{MAGI}    & Homo & \second{59.20}{0.39} & \res{69.89}{0.17} & \res{70.59}{0.27} & \res{11.02}{0.52} & \res{6.94}{0.19} & \res{50.43}{0.12} & \res{73.83}{0.21} & \res{44.25}{0.03} & \second{9.20}{0.03} & \res{52.89}{0.16} & \second{10.69}{0.30} & \second{56.33}{0.04} \\
&                          & Comp & \second{58.69}{0.42} & \res{67.47}{0.16} & \res{62.11}{0.12} & \res{11.46}{0.40} & \res{5.78}{0.14} & \res{43.19}{0.19} & \res{71.28}{0.22} & \res{38.79}{0.01} & \second{8.36}{0.02} & \res{38.53}{0.14} & \second{8.18}{0.15} & \second{50.15}{0.09} \\
\midrule

\multirow{10}{*}{\rotatebox{90}{Deep Joint}} 
& \multirow{2}{*}{DAEGC}   & Homo & \res{47.68}{1.58} & \res{65.16}{0.02} & \res{61.96}{0.03} & \res{11.91}{0.07} & \res{4.62}{0.02} & \res{44.05}{0.32} & \res{29.19}{1.11} & \res{30.63}{2.33} & \res{3.66}{0.24} & \res{10.06}{3.55} & \res{5.34}{0.31} & \res{25.27}{0.13} \\
&                          & Comp & \res{46.11}{1.84} & \res{62.11}{0.02} & \res{52.84}{0.03} & \res{11.05}{0.12} & \res{3.90}{0.03} & \res{36.32}{0.27} & \res{68.19}{1.84} & \res{27.53}{2.03} & \res{4.05}{0.17} & \res{8.44}{2.67} & \res{4.34}{0.27} & \res{32.92}{0.39} \\
\cmidrule{2-15}
& \multirow{2}{*}{DinkNet} & Homo & \res{55.95}{0.23} & \res{65.67}{0.04} & \res{62.09}{0.07} & \res{11.72}{0.07} & \res{6.26}{0.06} & \res{40.08}{0.74} & \res{55.99}{0.40} & \res{39.64}{0.06} & \res{3.98}{0.02} & \res{44.35}{0.19} & \res{6.18}{0.03} & \res{48.44}{0.04} \\
&                          & Comp & \res{55.18}{0.13} & \res{63.83}{0.09} & \res{53.27}{0.06} & \res{10.83}{0.05} & \res{7.15}{0.07} & \res{34.78}{0.46} & \res{53.96}{0.12} & \res{34.90}{0.05} & \res{3.85}{0.01} & \res{33.91}{0.18} & \res{4.91}{0.05} & \res{43.54}{0.07} \\
\cmidrule{2-15}
& \multirow{2}{*}{MinCut}  & Homo & \res{42.04}{1.90} & \res{64.38}{2.47} & \res{62.14}{2.23} & \res{7.13}{0.57}  & \res{8.17}{0.21} & \res{43.21}{1.28} & \res{49.17}{6.15} & \res{40.19}{0.14} & \res{6.06}{0.18} & \res{42.76}{0.91} & \res{7.01}{0.70} & -- \\
&                          & Comp & \res{39.64}{1.78} & \res{60.42}{2.31} & \res{52.54}{1.91} & \res{7.76}{0.17}  & \res{7.09}{0.26} & \res{35.53}{0.83} & \res{48.90}{0.89} & \res{36.42}{0.17} & \res{5.53}{0.17} & \res{30.79}{0.59} & \res{5.35}{0.41} & -- \\
\cmidrule{2-15}
& \multirow{2}{*}{DMoN}    & Homo & \res{44.95}{2.33} & \res{65.12}{2.78} & \res{63.54}{0.40} & \res{8.08}{0.97}  & \second{8.67}{0.09} & \res{43.04}{0.43} & \res{51.12}{0.36} & \res{41.45}{0.13} & \second{8.88}{0.09} & \res{41.64}{1.02} & \res{8.77}{0.48} & -- \\
&                          & Comp & \res{42.78}{2.27} & \res{60.52}{2.57} & \res{53.85}{0.34} & \res{7.16}{0.87}  & \best{7.23}{0.11} & \res{35.27}{0.19} & \res{50.21}{0.77} & \res{35.59}{0.11} & \second{8.00}{0.08} & \res{29.89}{0.72} & \res{6.54}{0.30} & -- \\
\cmidrule{2-15}
& \multirow{2}{*}{Neuromap}& Homo & \res{48.34}{2.67} & \res{63.20}{2.76} & \res{57.50}{0.17} & \res{7.51}{0.20}  & \res{8.47}{0.66} & \res{43.87}{0.72} & \res{49.70}{0.32} & \res{38.62}{0.20} & \res{8.81}{0.26} & \res{37.26}{1.83} & \res{8.00}{0.58} & -- \\
&                          & Comp & \res{45.70}{2.56} & \res{60.54}{1.96} & \res{55.67}{2.89} & \res{7.37}{0.15}  & \second{7.19}{0.51} & \res{37.96}{0.88} & \res{46.07}{0.25} & \second{39.68}{0.24} & \best{8.27}{0.23} & \res{32.66}{1.23} & \res{6.53}{0.46} & -- \\
\bottomrule
\end{tabular}
\end{adjustbox}
\end{table*}

\section{Dataset Descriptions}
\label{app:datasets}

In this section, we provide detailed descriptions of the 12 datasets curated for our benchmark. 
To comprehensively evaluate AGC algorithms, we selected datasets spanning diverse domains (Citation, Co-purchase, Social, Web) and feature modalities (Textual, Tabular). 
The datasets are categorized by scale, corresponding to the taxonomy presented in Table~\ref{tab:datasets}.
All datasets are accessed via the PyTorch Geometric~\cite{PyG} dataset interface\footnote{\url{https://pytorch-geometric.readthedocs.io/en/latest/modules/datasets.html}}, ensuring standardized preprocessing and reproducibility. 
The underlying datasets (e.g., OGB~\cite{OGB}, GraphLand~\cite{GraphLand}) are publicly available and distributed under permissive licenses (e.g., MIT), which allow for research use and redistribution.

\subsection{Tiny Scale Datasets ($N < 10^4$)}
\noindent\textbf{Cora}~\cite{GCN}. 
This is a standard citation network frequently used in graph clustering literature. Nodes represent machine learning papers, and edges represent citation links. The node attributes are sparse bag-of-words vectors indicating the presence of specific keywords in the paper. The clustering task involves grouping papers into seven distinct research topics.

\noindent\textbf{Photo}~\cite{Pitfall}. 
A segment of the Amazon co-purchase graph. Nodes represent goods (photo), and edges connect two products if they are frequently bought together. Node features are bag-of-words vectors extracted from product reviews. The goal is to cluster products into eight categories based on the product type.

\subsection{Small Scale Datasets ($10^4 \le N < 10^5$)}

\noindent\textbf{Physics}~\cite{Pitfall}. 
This is a co-authorship graph constructed from the Microsoft Academic Graph. Nodes are authors, and edges are established if two authors have co-authored a paper. Node features are bag-of-words representations of paper keywords. The authors are clustered into five research fields.

\noindent\textbf{HM}~\cite{GraphLand}. 
This dataset is derived from the H\&M Kaggle competition. 
The graph represents a co-purchasing network where nodes are products, and undirected edges connect products that are frequently bought by the same customers. The features are tabular, including product metadata (e.g., categorical features like color) and purchasing statistics (e.g., numerical features representing the proportion of purchases occurring on different weekdays). 
The task is to partition products into 21 distinct product categories.

\noindent\textbf{Flickr}~\cite{GraphSAINT}. 
An image-sharing social network dataset. 
Nodes represent images uploaded to Flickr, and edges reflect common metadata (e.g., same location, same gallery, or shared by the same user). 
Node features are 500-dimensional bag-of-words vectors based on user-provided tags. 
The clustering target is to group images into seven classes based on their tags.

\subsection{Medium Scale Datasets ($10^5 \le N < 10^6$)}

\noindent\textbf{ArXiv}~\cite{OGB}. 
A citation network representing Computer Science papers in arXiv. 
Nodes represent papers, and edges represent citations. 
Node features are 128-dimensional feature vectors obtained by averaging the embeddings of words in the paper's title and abstract. 
The task is to cluster papers into 40 subject areas.

\noindent\textbf{Reddit}~\cite{SAGE}. 
This dataset is constructed from Reddit posts made in the month of September 2014. 
Each node represents a post, and the ground-truth labels represent the community (subreddit) that the post belongs to. 
Nodes are connected based on common users commenting on both posts. 
Node features are 300-dimensional GloVe word vectors averaged over the content associated with the posts, including the title, comments, score, and number of comments.

\noindent\textbf{MAG}~\cite{OGB}. 
Extracted from the Microsoft Academic Graph, this is a heterogeneous network containing papers, authors, institutions, and fields of study. 
For the purpose of this benchmark, we utilize the homogeneous subgraph of papers connected by citation links. 
The node features are 128-dimensional Word2Vec embeddings of paper abstracts. 
The clustering task involves distinguishing between 349 venue categories (journals or conferences).

\subsection{Large Scale Datasets ($10^6 \le N < 10^8$)}

\noindent\textbf{Pokec}~\cite{GraphLand}. 
Based on data from the popular social network in Pokec. 
Nodes represent users, and directed edges connect users who have marked others as friends. 
The node features are tabular and derived from user profile information, including numerical features (e.g., profile completion proportion) and categorical indicators (e.g., whether specific profile fields are filled). 
The task is to predict which region a user belongs to, presenting an extreme multi-class clustering challenge with 183 classes and moderate heterophily.

\noindent\textbf{Products}~\cite{OGB}. 
A large-scale co-purchase network from Amazon. 
Nodes represent products, and edges connect products that are purchased together. 
Node features are generated from the product descriptions using a dimensionality-reduced bag-of-words approach. 
The clustering task is to group products into 47 top-level categories.

\noindent\textbf{WebTopic}~\cite{GraphLand}. 
This dataset represents a segment of the Internet (Web Graph) obtained from the Yandex search engine. 
Nodes are websites, and directed edges exist if a user followed a link from one website to another within a selected period. 
The features are tabular and highly heterogeneous, including numerical features (e.g., number of videos on the site) and categorical features (e.g., the website's zone, free hosting status). 
The goal is to cluster websites into 28 topics. 
This dataset is notable for its low homophily and complex feature distribution.

\subsection{Massive Scale Datasets ($N > 10^8$)}

\noindent\textbf{Papers100M}~\cite{OGB}. 
Currently one of the largest public graph benchmarks available. 
It is a citation network of approximately 111 million papers indexed by MAG. 
Directed edges represent citations. 
Node features are 128-dimensional embeddings averaged from the word embeddings of titles and abstracts. 
The task is to cluster papers into 172 subject areas. 
Note that approximately 1.5 million of nodes are labeled with one of arXiv's subject areas.

\section{Detailed Description of Evaluated Algorithms}
\label{app:baselines}

In this section, we provide detailed descriptions of the benchmark AGC methods. 
To facilitate a systematic comparison, we categorize these methods based on the Encode-Cluster-Optimize framework introduced in Section~\ref{sec:preliminaries}. 

\subsection{Traditional and Non-Parametric Methods}
These methods typically employ fixed or simplified encoding mechanisms and decouple the clustering process from representation learning.

\begin{itemize}[leftmargin=*]
    \item \textbf{KMeans~\cite{KMeans} \& Node2Vec~\cite{Node2Vec}:} We include these as fundamental representatives to assess the value of attribute-structure fusion. \textbf{KMeans} clusters raw node features (ignoring topology), while \textbf{Node2Vec} clusters structural embeddings learned via random walks (ignoring attributes).
    
    \item \textbf{SSGC~\cite{SSGC}:} 
    SSGC addresses the over-smoothing and scalability issues of deep GCNs. It utilizes a modified Markov Diffusion Kernel to derive a spectral graph convolution that balances low- and high-pass filter bands. In our framework, SSGC acts as a non-parametric encoder that aggregates features over multiple hops in a single step, followed by standard KMeans clustering. Its linearity allows for efficient processing, though it lacks the expressivity of deep non-linear encoders.

    \item \textbf{SAGSC~\cite{SAGSC}:} 
    SAGSC combines Laplacian smoothing with subspace clustering. It first smooths attributes over the graph to incorporate structural information. It then employs a self-expressive subspace clustering procedure that learns a factored coefficient matrix, projecting factors into a new space to generate a valid affinity matrix for spectral clustering. This method represents a hybrid approach where the encoding is a fixed smoothing operation, and the optimization targets self-expressiveness.
    
    \item \textbf{MS2CAG~\cite{MS2CAG}:} 
    MS2CAG improves upon subspace graph clustering by formulating a rank-constrained SVD problem. It introduces a linear-time optimization solver that avoids the explicit construction of the $N \times N$ self-expressive matrix, which is a common bottleneck in subspace clustering. Theoretical analysis links its objective to modularity maximization.
\end{itemize}

\subsection{Deep Decoupled Methods}
These methods utilize parametric Graph Neural Networks to learn node embeddings via self-supervised tasks, followed by a post-hoc application of a discrete clusterer (typically KMeans).

\begin{itemize}[leftmargin=*]
    \item \textbf{GAE~\cite{GAE}:} 
    A foundational generative method that uses a GNN encoder to parameterize the latent distribution of node embeddings. The objective is to reconstruct the adjacency matrix via a decoder. Clustering is performed on the learned representations.
    
    \item \textbf{DGI~\cite{DGI}:} 
    DGI relies on maximizing mutual information between local node patches and a global graph summary vector. It learns representations by contrasting true graph patches with corrupted counterparts (negative sampling). This contrastive objective encourages the encoder to capture global structural properties without explicit reconstruction.
    
    \item \textbf{CCASSG~\cite{CCASSG}:} 
    Unlike contrastive methods requiring negative pairs, CCASSG optimizes a feature-level objective inspired by Canonical Correlation Analysis. It generates two views via augmentation and minimizes the correlation between different feature dimensions while maximizing invariance across views. This decorrelation prevents collapse and removes the need for costly negative sampling.
    
    \item \textbf{S3GC~\cite{S3GC}:} 
    S3GC utilizes a contrastive learning framework specifically designed to enhance clusterability. It employs random walks to define positive neighborhoods and optimizes a contrastive loss that sharpens the decision boundaries between potential clusters. It effectively scales to massive-scale datasets.
    
    \item \textbf{NS4GC~\cite{NS4GC}:} 
    NS4GC argues that standard contrastive learning may neglect semantic node relations. It introduces a Reliable Node Similarity Matrix which guides the representation learning. The method aligns node neighbors and performs semantic-aware sparsification to ensure the learned embeddings preserve intrinsic semantic structures suited for clustering.
    
    \item \textbf{MAGI~\cite{MAGI}:} 
    MAGI bridges the gap between community detection and contrastive learning. It uses modularity maximization as a pretext task. Positive and negative pairs are defined not by random corruption, but by modularity-based communities, effectively using the GNN to maximize the modularity of the resulting partitions. This aligns the representation learning objective directly with the clustering goal.
\end{itemize}

\subsection{Deep Joint Methods}
These methods optimize the encoder and the cluster assignments simultaneously in an end-to-end fashion.

\begin{itemize}[leftmargin=*]
    \item \textbf{DAEGC~\cite{DAEGC}:} 
    DAEGC employs a Graph Attention Network as the encoder to capture neighbor importance. It combines a reconstruction loss with a clustering guidance loss (KL divergence between soft assignments and a target distribution). The self-training process iteratively refines the clusters and the embeddings.
    
    \item \textbf{DinkNet~\cite{DinkNet}:} 
    DinkNet is designed explicitly for scalability. It replaces the traditional KL-divergence loss with a Dilation (separating different clusters) and Shrink (compacting same-cluster nodes) loss. It initializes cluster centers as learnable parameters and uses an adversarial mechanism to optimize the distribution.
    
    \item \textbf{MinCut~\cite{MinCut}:} 
    This method provides a continuous relaxation of the normalized min-cut problem. A GNN projects nodes directly into a soft cluster assignment matrix. The loss function minimizes the cut value (edge connections between clusters) while enforcing orthogonality to prevent trivial solutions (all nodes in one cluster).
    
    \item \textbf{DMoN~\cite{DMoN}:} 
    DMoN optimizes the spectral modularity objective. It essentially acts as a differentiable pooling operator that coarsens the graph. Unlike MinCut, it does not require orthogonality constraints because the modularity metric naturally penalizes trivial partitions.
    
    \item \textbf{Neuromap~\cite{Neuromap}:} 
    Neuromap creates a differentiable formulation of the Map Equation, an information-theoretic measure for community detection based on flow dynamics. It optimizes the code length required to describe random walks on the graph, naturally balancing cluster internal density with external separation without requiring a pre-specified number of clusters, though we fix $K$ for consistent benchmarking.
\end{itemize}

\section{Implementation Details}
\label{app:implementation_details}

\vpara{Hardware Environment.} 
To ensure fair comparisons and demonstrate the accessibility of our benchmark to standard research laboratories, all experiments are conducted on a Linux workstation equipped with an Intel(R) Xeon(R) Platinum 8163 CPU @ 2.50GHz, 480GB RAM, and a single NVIDIA Tesla V100 (32GB) GPU. 
By restricting evaluation to a single GPU, we stress-test the memory efficiency of the proposed mini-batch implementations, verifying that our scalable solutions are relevant to industrial and academic settings without reliance on massive multi-GPU clusters.

\vpara{Reproducibility and Hyperparameters.} 
To resolve the reproducibility crisis often observed in clustering research, we do not hard-code parameters. Instead, all specific hyperparameters (e.g., learning rate, encoder depth, hidden dimensions) are strictly managed via configuration files (YAML) located in each method's directory (e.g., \texttt{benchmark/DMoN/train.conf.yaml}). 
The code and resources are publicly available via \href{https://github.com/Cloudy1225/PyAGC}{\texttt{GitHub}}, \href{https://pypi.org/project/pyagc}{\texttt{PyPI}}, and \href{https://pyagc.readthedocs.io}{\texttt{Docs}}.

\vpara{Training Protocol.} 
All deep learning methods are optimized using the Adam optimizer. 
To ensure statistical reliability and mitigate the impact of random initialization, we report the mean and standard deviation of all metrics across 5 independent runs with different random seeds.
For all datasets except \texttt{Papers100M}, models are trained until convergence or for a fixed number of epochs.

Regarding the massive \texttt{Papers100M} dataset, due to its prohibitive computational cost, we restrict the training of deep methods to a single epoch. This design is consistent with established practice at this scale: prior works such as GGD~\cite{GGD} and DinkNet~\cite{DinkNet} also adopt 1-epoch training on \texttt{Papers100M}, while MAGI~\cite{MAGC} caps runtime at 6 hours, which corresponds to approximately one epoch in our setting. 
Beyond convention, this constraint reflects a realistic industrial scenario where models are retrained under strict time budgets. In such settings, rapid convergence is a desirable property, and methods that achieve strong performance within a single epoch (e.g., CCASSG: 1.29h, NMI=53.82\% on \texttt{Papers100M}) are inherently more practical. Our benchmark thus explicitly evaluates this trade-off between efficiency and performance, treating it as an important aspect of real-world applicability.

\begin{table*}[h]
\centering
\caption{
Encoder ablation study on four representative datasets. 
We report NMI, ACC, and Modularity (Mod) in percentage form as mean $\pm$ standard deviation.
For each method and dataset, the \colorbox{colorBest}{\textbf{best}} and \colorbox{colorSecond}{second-best} encoder results are highlighted for each metric.
}
\label{tab:encoder_ablation}
\vspace{-0.1cm}
\scriptsize
\setlength{\tabcolsep}{2.6pt}
\renewcommand{\arraystretch}{1.08}
\begin{adjustbox}{width=\textwidth, center}
\begin{tabular}{ll ccc ccc ccc ccc}
\toprule
\multirow{2}{*}{\textbf{Method}} 
& \multirow{2}{*}{\textbf{Encoder}} 
& \multicolumn{3}{c}{\textbf{Cora}} 
& \multicolumn{3}{c}{\textbf{HM}} 
& \multicolumn{3}{c}{\textbf{ArXiv}} 
& \multicolumn{3}{c}{\textbf{Pokec}} \\
\cmidrule(lr){3-5} \cmidrule(lr){6-8} \cmidrule(lr){9-11} \cmidrule(lr){12-14}
& & \textbf{NMI} & \textbf{ACC} & \textbf{Mod}
  & \textbf{NMI} & \textbf{ACC} & \textbf{Mod}
  & \textbf{NMI} & \textbf{ACC} & \textbf{Mod}
  & \textbf{NMI} & \textbf{ACC} & \textbf{Mod} \\
\midrule

\multirow{5}{*}{\textbf{DGI}} 
& \texttt{GCN}   
& \secondenc{56.22$\pm$0.87} & 72.22$\pm$2.92 & 71.49$\pm$0.28 
& \bestenc{11.67$\pm$0.06} & \bestenc{15.78$\pm$0.32} & \bestenc{5.53$\pm$0.10} 
& \bestenc{42.29$\pm$0.08} & \bestenc{30.49$\pm$0.64} & 54.30$\pm$0.50 
& 4.20$\pm$0.01 & 2.23$\pm$0.02 & 6.65$\pm$0.04 \\

& \texttt{SAGE}  
& 52.09$\pm$0.29 & 70.01$\pm$0.52 & 69.82$\pm$0.13 
& 8.76$\pm$0.29 & 13.42$\pm$0.37 & 3.07$\pm$0.08 
& 39.28$\pm$0.13 & 26.83$\pm$0.51 & 48.23$\pm$0.57 
& 2.28$\pm$0.01 & 1.64$\pm$0.01 & 2.15$\pm$0.02 \\

& \texttt{GIN}   
& 32.72$\pm$1.66 & 41.82$\pm$3.16 & 32.54$\pm$5.09 
& \secondenc{8.92$\pm$0.03} & \secondenc{14.15$\pm$0.18} & \secondenc{3.29$\pm$1.14} 
& 35.67$\pm$0.30 & \secondenc{30.04$\pm$1.19} & 47.20$\pm$0.73 
& 3.83$\pm$0.03 & 2.36$\pm$0.02 & 7.20$\pm$0.03 \\

& \texttt{GAT}   
& \bestenc{56.23$\pm$0.41} & \bestenc{72.93$\pm$0.82} & \secondenc{72.10$\pm$0.21} 
& 7.24$\pm$0.13 & 13.41$\pm$0.27 & -0.19$\pm$0.76 
& 41.07$\pm$0.07 & 28.94$\pm$0.17 & \bestenc{56.06$\pm$0.08} 
& \bestenc{5.14$\pm$0.02} & \bestenc{2.65$\pm$0.02} & \bestenc{8.99$\pm$0.05} \\

& \texttt{GATv2} 
& 56.20$\pm$0.51 & \secondenc{72.90$\pm$0.79} & \bestenc{72.13$\pm$0.20} 
& 7.24$\pm$0.13 & 13.42$\pm$0.28 & -0.15$\pm$0.78 
& \secondenc{41.12$\pm$0.10} & 28.46$\pm$0.82 & \secondenc{55.60$\pm$0.16} 
& \secondenc{5.06$\pm$0.02} & \secondenc{2.64$\pm$0.02} & \secondenc{8.51$\pm$0.08} \\
\midrule

\multirow{5}{*}{\textbf{NS4GC}} 
& \texttt{GCN}   
& \bestenc{59.40$\pm$0.48} & \bestenc{74.92$\pm$0.59} & \secondenc{75.08$\pm$0.25} 
& 10.93$\pm$0.12 & 14.73$\pm$0.09 & -0.57$\pm$0.04 
& \secondenc{48.28$\pm$0.16} & 36.87$\pm$0.84 & 61.61$\pm$0.36 
& \secondenc{7.19$\pm$0.04} & 3.19$\pm$0.01 & 15.98$\pm$0.09 \\

& \texttt{SAGE}  
& 53.58$\pm$2.14 & 71.02$\pm$1.70 & 74.34$\pm$0.51 
& \bestenc{15.28$\pm$0.17} & \bestenc{18.40$\pm$0.42} & 5.11$\pm$0.19 
& \bestenc{48.39$\pm$0.24} & 39.01$\pm$1.07 & 62.37$\pm$0.27 
& \bestenc{12.00$\pm$0.02} & \bestenc{4.88$\pm$0.02} & \bestenc{23.81$\pm$0.02} \\

& \texttt{GIN}   
& 49.63$\pm$1.75 & 64.31$\pm$3.14 & 74.80$\pm$0.58 
& 8.42$\pm$0.06 & 14.90$\pm$0.18 & 3.25$\pm$0.53 
& 40.14$\pm$0.21 & 29.37$\pm$0.40 & 31.60$\pm$0.45 
& 7.00$\pm$0.01 & \secondenc{3.69$\pm$0.04} & 16.33$\pm$0.24 \\

& \texttt{GAT}   
& 57.51$\pm$1.79 & 73.77$\pm$2.41 & \bestenc{75.12$\pm$0.11} 
& \secondenc{12.05$\pm$0.12} & \secondenc{17.57$\pm$0.48} & \secondenc{10.91$\pm$0.95} 
& 48.04$\pm$0.16 & \bestenc{40.02$\pm$0.65} & \secondenc{63.86$\pm$0.24} 
& 6.09$\pm$0.01 & 3.03$\pm$0.03 & 16.72$\pm$0.25 \\

& \texttt{GATv2} 
& \secondenc{57.94$\pm$0.07} & \secondenc{74.50$\pm$0.21} & 74.60$\pm$0.28 
& 11.96$\pm$0.04 & 17.20$\pm$0.12 & \bestenc{11.55$\pm$0.09} 
& 47.94$\pm$0.13 & \secondenc{39.73$\pm$0.56} & \bestenc{63.93$\pm$0.22} 
& 6.08$\pm$0.02 & 3.04$\pm$0.01 & \secondenc{16.87$\pm$0.07} \\
\midrule

\multirow{5}{*}{\textbf{DMoN}} 
& \texttt{GCN}   
& 43.84$\pm$2.29 & 52.99$\pm$4.70 & \secondenc{71.81$\pm$1.40} 
& 7.59$\pm$0.92 & 12.32$\pm$0.58 & \bestenc{12.54$\pm$0.25} 
& 38.77$\pm$0.28 & \secondenc{25.89$\pm$1.12} & 61.14$\pm$1.19 
& 8.42$\pm$0.08 & 4.40$\pm$0.07 & \secondenc{22.67$\pm$0.28} \\

& \texttt{SAGE}  
& \bestenc{45.99$\pm$2.09} & \bestenc{56.23$\pm$2.93} & \bestenc{73.97$\pm$0.32} 
& \bestenc{11.86$\pm$0.19} & 15.94$\pm$0.40 & \secondenc{8.58$\pm$0.44} 
& 39.61$\pm$0.32 & 24.35$\pm$0.86 & 65.35$\pm$0.27 
& \secondenc{9.62$\pm$0.20} & \secondenc{4.74$\pm$0.16} & 21.47$\pm$0.42 \\

& \texttt{GIN}   
& 40.69$\pm$1.82 & 50.85$\pm$3.04 & 71.03$\pm$1.31 
& 1.48$\pm$1.03 & \bestenc{17.97$\pm$0.30} & 0.30$\pm$0.55 
& 21.99$\pm$2.61 & 20.30$\pm$1.59 & 40.20$\pm$7.63 
& \bestenc{11.84$\pm$0.05} & \bestenc{6.30$\pm$0.21} & \bestenc{28.72$\pm$0.36} \\

& \texttt{GAT}   
& \secondenc{43.87$\pm$6.04} & \secondenc{53.67$\pm$4.67} & 69.68$\pm$4.59 
& \secondenc{8.72$\pm$0.83} & \secondenc{16.27$\pm$1.16} & 5.42$\pm$1.14 
& \bestenc{40.86$\pm$0.33} & \bestenc{26.06$\pm$0.72} & \bestenc{65.48$\pm$0.18} 
& 7.38$\pm$0.42 & 3.82$\pm$0.18 & 16.44$\pm$0.83 \\

& \texttt{GATv2} 
& 42.96$\pm$5.73 & 53.23$\pm$4.24 & 69.64$\pm$4.62 
& \secondenc{8.72$\pm$0.83} & 16.26$\pm$1.16 & 5.42$\pm$1.14 
& \secondenc{40.79$\pm$0.34} & 25.82$\pm$0.65 & \secondenc{65.46$\pm$0.25} 
& 7.38$\pm$0.43 & 3.82$\pm$0.18 & 16.44$\pm$0.83 \\
\midrule

\multirow{5}{*}{\textbf{Neuromap}} 
& \texttt{GCN}   
& \bestenc{46.98$\pm$2.61} & 56.96$\pm$3.98 & \bestenc{74.02$\pm$0.95} 
& \secondenc{7.44$\pm$0.17} & 12.73$\pm$0.25 & \bestenc{6.29$\pm$0.17} 
& \bestenc{40.70$\pm$0.80} & 33.31$\pm$1.07 & 64.54$\pm$0.66 
& 8.53$\pm$0.25 & 5.12$\pm$0.21 & 31.71$\pm$0.56 \\

& \texttt{SAGE}  
& 42.60$\pm$3.71 & 53.47$\pm$4.35 & \secondenc{72.64$\pm$1.25} 
& \bestenc{9.31$\pm$0.12} & 13.86$\pm$0.17 & \secondenc{5.11$\pm$0.14} 
& 40.29$\pm$0.79 & \bestenc{36.84$\pm$2.36} & \bestenc{66.47$\pm$0.74} 
& \secondenc{10.94$\pm$0.14} & \bestenc{6.67$\pm$0.25} & \secondenc{33.32$\pm$0.52} \\

& \texttt{GIN}   
& 39.03$\pm$2.14 & 49.00$\pm$3.01 & 70.88$\pm$0.85 
& 3.16$\pm$0.80 & \bestenc{15.17$\pm$0.49} & 4.35$\pm$2.24 
& 27.76$\pm$1.62 & 28.71$\pm$2.79 & 54.00$\pm$2.14 
& \bestenc{11.16$\pm$0.16} & \secondenc{6.51$\pm$0.19} & \bestenc{38.61$\pm$0.37} \\

& \texttt{GAT}   
& \secondenc{45.27$\pm$1.80} & \bestenc{58.97$\pm$3.11} & 71.97$\pm$0.91 
& 6.35$\pm$0.30 & \secondenc{14.06$\pm$0.70} & 4.25$\pm$0.85 
& 40.44$\pm$0.68 & 36.12$\pm$0.82 & 65.16$\pm$0.71 
& 2.90$\pm$0.04 & 2.50$\pm$0.05 & 10.79$\pm$0.23 \\

& \texttt{GATv2} 
& 44.62$\pm$1.95 & \secondenc{58.06$\pm$3.18} & 72.17$\pm$1.08 
& 6.37$\pm$0.32 & 13.98$\pm$0.71 & 4.11$\pm$0.99 
& \secondenc{40.50$\pm$0.75} & \secondenc{36.28$\pm$1.26} & \secondenc{65.30$\pm$0.44} 
& 2.90$\pm$0.04 & 2.50$\pm$0.05 & 10.79$\pm$0.24 \\
\bottomrule
\end{tabular}
\end{adjustbox}
\vspace{-0.2cm}
\end{table*}

\section{Ablation Study on Graph Encoders}
\label{app:encoder_ablation}

To assess the impact of encoder choice on clustering performance, we conduct a systematic ablation study across four representative AGC methods (two deep decoupled methods DGI and NS4GC; two deep joint methods DMoN and Neuromap). 
Non-parametric methods (e.g., SSGC, SAGSC, MS2CAG) are excluded from this ablation, as they rely on fixed graph encoders. 
Each method is instantiated with five encoder backbones: \texttt{GCN}~\cite{GCN}, \texttt{SAGE}~\cite{SAGE}, \texttt{GIN}~\cite{GIN}, \texttt{GAT}~\cite{GAT}, and \texttt{GATv2}~\cite{GATv2}. 
Crucially, all encoders follow the tuned implementations from~\cite{TunedGNN}, which incorporate normalization, residual connections, and other engineering improvements that yield competitive performance under both \emph{homophilous} and \emph{heterophilous} settings. We evaluate on four datasets that represent the key axes of diversity in our benchmark: \texttt{Cora} (tiny, high-homophily, textual), \texttt{HM} (small, low-homophily, tabular), \texttt{ArXiv} (medium, moderate-homophily, textual), and \texttt{Pokec} (large, moderate-heterophily, tabular). Full results are reported in Tables~\ref{tab:encoder_ablation} and~\ref{tab:encoder_ablation_ari_f1_cond}.

\vpara{Observations.}
The most salient finding is that no encoder achieves consistently superior performance across all methods, datasets, and metrics. For instance, within DGI, \texttt{GAT} and \texttt{GATv2} attain the highest modularity on \texttt{Cora} and \texttt{Pokec}, yet \texttt{GCN} leads on \texttt{HM} by a substantial margin in NMI and ACC. Similarly, for NS4GC, \texttt{SAGE} yields the best supervised metrics on \texttt{HM} and \texttt{Pokec}, while \texttt{GAT}/\texttt{GATv2} achieve the strongest structural quality (Modularity, Conductance) on \texttt{ArXiv}. This underscores that encoder selection constitutes a non-trivial, dataset-dependent design decision, and practitioners should not default to a canonical choice such as GCN without empirical validation.

On the high-homophily \texttt{Cora} dataset, attention-based encoders (\texttt{GAT}, \texttt{GATv2}) and \texttt{GCN} perform comparably and generally outperform \texttt{GIN}, which exhibits substantially higher variance and lower absolute performance. This aligns with the well-known limitation of sum-aggregation in GIN when the clustering signal is primarily driven by smooth feature propagation. In contrast, on the heterophilous \texttt{HM} ($\mathcal{H}_e = 0.16$) and \texttt{Pokec} ($\mathcal{H}_e = 0.43$) datasets, the relative rankings shift considerably: \texttt{SAGE}'s mean-aggregation proves more robust to boundary noise in several configurations (e.g., NS4GC on \texttt{HM} and \texttt{Pokec}), while \texttt{GIN} occasionally excels on unsupervised structural metrics such as Conductance (e.g., DMoN and Neuromap on \texttt{Pokec}), suggesting it may discover topologically denser partitions even when supervised alignment is poor.

\texttt{GAT} and \texttt{GATv2} produce nearly indistinguishable results, with differences consistently within one standard deviation. This indicates that the dynamic attention mechanism of GATv2~\cite{GATv2} provides negligible additional expressivity in the AGC context, where the primary bottleneck is community structure rather than fine-grained pairwise attention. For practitioners seeking to reduce architectural search cost, \texttt{GAT} serves as a sufficient proxy.

We observe cases where the encoder optimal for supervised metrics (NMI, ACC, ARI, F1) differs from the one optimal for structural quality (Modularity, Conductance). For example, in DMoN on \texttt{Pokec}, \texttt{GIN} achieves the best Conductance (69.57\%) and second-best Modularity, yet its NMI (11.84\%) does not lead all encoders. Conversely, \texttt{GCN} attains competitive NMI but inferior structural quality. This reinforces the central argument of our holistic evaluation protocol (\S\ref{sec:evaluation_protocol}): in label-scarce industrial settings, encoder selection driven purely by supervised metrics may be misleading, and structural metrics provide a complementary, indispensable signal.

Deep joint methods (DMoN, Neuromap) exhibit higher encoder sensitivity than the decoupled method DGI, as evidenced by larger performance gaps between the best and worst encoders within a given dataset. For instance, Neuromap on \texttt{Pokec} spans a Conductance range of roughly 57\%--87\% across encoders, while DGI spans a narrower 90\%--97\% range. This suggests that the end-to-end coupling between encoder and clustering objective amplifies the effect of encoder inductive bias, making architecture search more critical for joint methods.

Absent dataset-specific tuning, \texttt{SAGE} represents a reliable default across the heterophilous industrial graphs in our benchmark (\texttt{HM}, \texttt{Pokec}), while \texttt{GCN} and \texttt{GAT} remain competitive on homophilous citation graphs. We recommend practitioners run a lightweight encoder sweep, enabled by the modular design of \texttt{PyAGC}, rather than committing to a single backbone, as the performance differential between the best and worst encoder can exceed 10 NMI points on industrial datasets.

\begin{table*}[h]
\centering
\caption{
Encoder ablation study on four representative datasets.
We report ARI, F1, and Conductance (Cond) in percentage form as mean $\pm$ standard deviation.
For ARI and F1, higher is better; for Cond, lower is better.
For each method and dataset, the \colorbox{colorBest}{\textbf{best}} and \colorbox{colorSecond}{second-best} encoder results are highlighted for each metric.
}
\label{tab:encoder_ablation_ari_f1_cond}
\vspace{-0.1cm}
\scriptsize
\setlength{\tabcolsep}{2.6pt}
\renewcommand{\arraystretch}{1.08}
\begin{adjustbox}{width=\textwidth, center}
\begin{tabular}{ll ccc ccc ccc ccc}
\toprule
\multirow{2}{*}{\textbf{Method}} 
& \multirow{2}{*}{\textbf{Encoder}} 
& \multicolumn{3}{c}{\textbf{Cora}} 
& \multicolumn{3}{c}{\textbf{HM}} 
& \multicolumn{3}{c}{\textbf{ArXiv}} 
& \multicolumn{3}{c}{\textbf{Pokec}} \\
\cmidrule(lr){3-5} \cmidrule(lr){6-8} \cmidrule(lr){9-11} \cmidrule(lr){12-14}
& & \textbf{ARI} & \textbf{F1} & \textbf{Cond$\downarrow$}
  & \textbf{ARI} & \textbf{F1} & \textbf{Cond$\downarrow$}
  & \textbf{ARI} & \textbf{F1} & \textbf{Cond$\downarrow$}
  & \textbf{ARI} & \textbf{F1} & \textbf{Cond$\downarrow$} \\
\midrule

\multirow{5}{*}{\textbf{DGI}} 
& \texttt{GCN} 
& \secondenc{52.03$\pm$3.22} & 70.07$\pm$2.56 & 12.43$\pm$0.78 
& \bestenc{3.11$\pm$0.06} & \bestenc{14.20$\pm$0.23} & \secondenc{67.22$\pm$1.10} 
& \secondenc{22.77$\pm$0.28} & \bestenc{21.47$\pm$0.30} & 39.15$\pm$0.69 
& 0.32$\pm$0.00 & \secondenc{1.59$\pm$0.01} & 92.40$\pm$0.03 \\

& \texttt{SAGE} 
& 46.53$\pm$0.73 & 68.91$\pm$0.42 & 14.97$\pm$0.16 
& \secondenc{2.43$\pm$0.08} & 12.00$\pm$0.38 & 90.49$\pm$0.28 
& 19.64$\pm$0.26 & 18.99$\pm$0.22 & 45.56$\pm$0.58 
& 0.15$\pm$0.00 & 1.14$\pm$0.01 & 96.63$\pm$0.04 \\

& \texttt{GIN} 
& 21.18$\pm$1.68 & 41.52$\pm$4.82 & 48.89$\pm$4.12 
& 2.27$\pm$0.03 & \secondenc{12.04$\pm$0.11} & \bestenc{66.44$\pm$1.29} 
& \bestenc{24.34$\pm$1.31} & 18.63$\pm$0.71 & 40.20$\pm$1.82 
& 0.32$\pm$0.01 & 1.52$\pm$0.01 & 91.17$\pm$0.04 \\

& \texttt{GAT} 
& \bestenc{52.04$\pm$1.61} & \bestenc{71.29$\pm$0.61} & \secondenc{11.88$\pm$0.52} 
& 1.71$\pm$0.06 & 11.68$\pm$0.47 & 70.76$\pm$2.39 
& 22.37$\pm$0.12 & 20.10$\pm$0.11 & \bestenc{36.39$\pm$0.16} 
& \bestenc{0.50$\pm$0.00} & \bestenc{1.84$\pm$0.01} & \bestenc{90.03$\pm$0.06} \\

& \texttt{GATv2} 
& 52.02$\pm$1.63 & \secondenc{71.24$\pm$0.55} & \bestenc{11.83$\pm$0.52} 
& 1.71$\pm$0.06 & 11.68$\pm$0.47 & 70.64$\pm$2.51 
& 21.88$\pm$0.43 & \secondenc{20.21$\pm$0.22} & \secondenc{37.30$\pm$0.35} 
& \secondenc{0.49$\pm$0.00} & \bestenc{1.84$\pm$0.01} & \secondenc{90.53$\pm$0.09} \\
\midrule

\multirow{5}{*}{\textbf{NS4GC}} 
& \texttt{GCN} 
& \bestenc{56.63$\pm$1.20} & \bestenc{72.81$\pm$0.38} & \secondenc{8.49$\pm$0.12} 
& \secondenc{2.72$\pm$0.03} & \secondenc{14.03$\pm$0.11} & \bestenc{53.24$\pm$0.58} 
& 25.83$\pm$0.85 & \bestenc{27.32$\pm$0.42} & 31.30$\pm$0.58 
& 0.74$\pm$0.01 & 2.33$\pm$0.02 & 82.72$\pm$0.10 \\

& \texttt{SAGE} 
& 49.49$\pm$2.36 & 69.24$\pm$1.85 & 9.00$\pm$0.52 
& \bestenc{4.72$\pm$0.10} & \bestenc{17.23$\pm$0.28} & 85.63$\pm$0.38 
& \secondenc{31.84$\pm$2.30} & 26.67$\pm$0.57 & 29.25$\pm$0.67 
& \bestenc{1.58$\pm$0.03} & \bestenc{3.34$\pm$0.01} & \bestenc{75.05$\pm$0.03} \\

& \texttt{GIN} 
& 40.04$\pm$3.83 & 64.58$\pm$2.59 & 9.69$\pm$0.46 
& 1.93$\pm$0.05 & 12.33$\pm$0.18 & 71.48$\pm$1.32 
& 19.42$\pm$0.67 & 22.41$\pm$0.22 & 59.06$\pm$0.80 
& \secondenc{1.03$\pm$0.02} & \secondenc{2.48$\pm$0.03} & 82.22$\pm$0.28 \\

& \texttt{GAT} 
& \secondenc{54.52$\pm$3.31} & 71.86$\pm$1.85 & \bestenc{8.35$\pm$0.26} 
& 2.69$\pm$0.07 & 13.87$\pm$0.36 & 57.60$\pm$2.37 
& \bestenc{33.82$\pm$1.37} & \secondenc{26.69$\pm$0.33} & \secondenc{27.02$\pm$0.50} 
& 0.65$\pm$0.01 & 2.16$\pm$0.02 & 81.89$\pm$0.28 \\

& \texttt{GATv2} 
& 54.48$\pm$0.23 & \secondenc{72.54$\pm$0.30} & 8.82$\pm$0.08 
& 2.68$\pm$0.05 & 13.51$\pm$0.07 & \secondenc{56.28$\pm$0.39} 
& \bestenc{33.82$\pm$1.28} & 26.59$\pm$0.39 & \bestenc{26.94$\pm$0.49} 
& 0.65$\pm$0.00 & 2.16$\pm$0.01 & \secondenc{81.72$\pm$0.09} \\
\midrule

\multirow{5}{*}{\textbf{DMoN}} 
& \texttt{GCN} 
& 33.35$\pm$3.35 & \secondenc{49.57$\pm$3.84} & \secondenc{12.63$\pm$0.86} 
& 2.05$\pm$0.28 & 10.58$\pm$0.44 & 66.99$\pm$0.99 
& 15.44$\pm$0.37 & \bestenc{20.37$\pm$0.61} & 33.53$\pm$0.88 
& 1.17$\pm$0.04 & 3.04$\pm$0.04 & \secondenc{76.06$\pm$0.30} \\

& \texttt{SAGE} 
& \bestenc{35.47$\pm$2.42} & \bestenc{53.53$\pm$2.96} & \bestenc{11.02$\pm$0.47} 
& \bestenc{3.49$\pm$0.10} & \bestenc{14.34$\pm$0.35} & 80.66$\pm$1.65 
& 15.26$\pm$0.24 & 19.82$\pm$0.82 & 30.64$\pm$0.30 
& \secondenc{1.33$\pm$0.06} & \secondenc{3.31$\pm$0.09} & 77.54$\pm$0.42 \\

& \texttt{GIN} 
& 30.74$\pm$1.74 & 47.97$\pm$3.69 & 12.70$\pm$1.05 
& -0.04$\pm$0.04 & 2.52$\pm$0.84 & \bestenc{0.30$\pm$0.47} 
& 6.88$\pm$1.71 & 6.64$\pm$1.27 & 40.29$\pm$9.97 
& \bestenc{1.66$\pm$0.10} & \bestenc{3.51$\pm$0.07} & \bestenc{69.57$\pm$0.39} \\

& \texttt{GAT} 
& \secondenc{34.93$\pm$5.06} & 46.80$\pm$4.14 & 13.62$\pm$4.78 
& \secondenc{2.97$\pm$0.50} & \secondenc{10.98$\pm$0.78} & \secondenc{65.46$\pm$16.08} 
& \secondenc{16.43$\pm$0.35} & \secondenc{20.19$\pm$0.74} & \secondenc{30.43$\pm$0.14} 
& 0.98$\pm$0.06 & 2.72$\pm$0.16 & 82.65$\pm$0.80 \\

& \texttt{GATv2} 
& 33.64$\pm$4.43 & 46.93$\pm$4.25 & 13.81$\pm$4.75 
& \secondenc{2.97$\pm$0.50} & \secondenc{10.98$\pm$0.78} & \secondenc{65.46$\pm$16.08} 
& \bestenc{16.51$\pm$0.17} & 19.92$\pm$0.62 & \bestenc{30.41$\pm$0.32} 
& 0.98$\pm$0.06 & 2.72$\pm$0.17 & 82.64$\pm$0.80 \\
\midrule

\multirow{5}{*}{\textbf{Neuromap}} 
& \texttt{GCN} 
& \bestenc{36.69$\pm$3.25} & \bestenc{54.69$\pm$3.95} & \bestenc{10.96$\pm$0.93} 
& 1.45$\pm$0.07 & \secondenc{9.37$\pm$0.18} & 20.28$\pm$0.45 
& 21.92$\pm$1.08 & \bestenc{21.29$\pm$0.84} & 27.76$\pm$1.09 
& 1.84$\pm$0.11 & \secondenc{2.62$\pm$0.14} & 65.39$\pm$0.65 \\

& \texttt{SAGE} 
& 32.42$\pm$4.54 & 50.77$\pm$4.30 & 12.14$\pm$1.00 
& \bestenc{2.04$\pm$0.06} & \bestenc{11.36$\pm$0.13} & 21.50$\pm$1.91 
& \bestenc{27.01$\pm$1.23} & 17.17$\pm$1.96 & \bestenc{23.11$\pm$1.02} 
& \bestenc{2.56$\pm$0.12} & \bestenc{3.25$\pm$0.10} & \secondenc{64.38$\pm$0.60} \\

& \texttt{GIN} 
& 28.74$\pm$1.42 & 46.36$\pm$4.01 & 12.81$\pm$0.64 
& 0.68$\pm$0.17 & 4.81$\pm$1.25 & 17.25$\pm$8.32 
& 18.12$\pm$1.93 & 10.56$\pm$1.02 & 30.67$\pm$3.43 
& \secondenc{2.50$\pm$0.08} & 2.50$\pm$0.12 & \bestenc{57.52$\pm$0.50} \\

& \texttt{GAT} 
& \secondenc{36.29$\pm$2.84} & \secondenc{53.87$\pm$3.47} & 11.81$\pm$0.41 
& \secondenc{1.49$\pm$0.12} & 8.22$\pm$0.78 & \secondenc{15.16$\pm$2.06} 
& 25.36$\pm$0.80 & \secondenc{19.39$\pm$1.04} & 25.52$\pm$0.98 
& 0.53$\pm$0.01 & 1.34$\pm$0.04 & 87.48$\pm$0.28 \\

& \texttt{GATv2} 
& 34.87$\pm$2.98 & 53.74$\pm$3.47 & \secondenc{11.78$\pm$0.41} 
& 1.48$\pm$0.09 & 8.28$\pm$0.77 & \bestenc{15.07$\pm$2.12} 
& \secondenc{25.52$\pm$1.67} & 19.37$\pm$0.83 & \secondenc{25.28$\pm$1.15} 
& 0.53$\pm$0.01 & 1.34$\pm$0.04 & 87.49$\pm$0.30 \\
\bottomrule
\end{tabular}
\end{adjustbox}
\vspace{-0.2cm}
\end{table*}

\section{Ablation Study on Neural Tabular Encoding}
\label{app:tabular_encoder}

A key challenge in industrial graph clustering is the effective encoding of heterogeneous tabular features. 
Following GraphLand~\cite{GraphLand}, our default preprocessing protocol, denoted as \textsc{Base}, transforms categorical features by one-hot encoding and maps numerical features to a standard normal distribution using quantile transformation.
While this protocol is simple and robust, recent progress in tabular deep learning suggests that learnable feature tokenization may better preserve field-specific semantics~\cite{FT-Transformer}.
Therefore, we further investigate whether a \emph{neural tabular encoder} can improve AGC performance on industrial tabular graphs (\texttt{HM}, \texttt{Pokec}, \texttt{WebTopic}).

\vpara{Neural Tabular Encoding.}
The proposed encoder replaces static feature preprocessing with learnable transformations.
For categorical features, an embedding lookup layer maps each discrete token to a dense vector.
For numerical features, a per-feature linear projection is applied to each raw scalar, and the resulting embeddings are concatenated.
The unified feature representation is then passed through an MLP projection layer before being fed to the GNN backbone.
This design closely follows the feature tokenizer in FT-Transformer~\cite{FT-Transformer}, adapted to the graph setting.

\vpara{Experimental Setup.}
We conduct ablation experiments on four representative AGC methods (two deep decoupled methods DGI and NS4GC; two deep joint methods DMoN and Neuromap).
Non-parametric methods (e.g., SSGC, SAGSC, MS2CAG) are excluded from this ablation, as they rely on fixed graph filters without learnable node embeddings and thus cannot incorporate a neural tabular encoder.
Note that DMoN and Neuromap encounter out-of-memory (OOM) errors on \texttt{WebTopic} under \textsc{+Tabular} due to the additional parameters introduced by the neural encoder.
All other settings follow the main benchmark protocol (Section~\ref{sec:setup}).
Results are reported in Table~\ref{tab:tabular_encoder_ablation}.

\begin{table*}[t]
\centering
\caption{
Ablation study of neural tabular encoding on industrial tabular attributed graphs.
\textsc{Base} denotes the classic tabular preprocessing protocol.
\textsc{+Tabular} introduces a neural tabular encoder.
The better result between \textsc{Base} and \textsc{+Tabular} is highlighted for each method and metric.
}
\label{tab:tabular_encoder_ablation}
\vspace{-0.1cm}
\scriptsize
\setlength{\tabcolsep}{4.0pt}
\renewcommand{\arraystretch}{1.08}
\begin{adjustbox}{width=\textwidth, center}
\begin{tabular}{lll cccccc}
\toprule
\textbf{Dataset} 
& \textbf{Method} 
& \textbf{Variant} 
& \textbf{NMI$\uparrow$} 
& \textbf{ARI$\uparrow$} 
& \textbf{ACC$\uparrow$} 
& \textbf{F1$\uparrow$} 
& \textbf{Mod$\uparrow$} 
& \textbf{Cond$\downarrow$} \\
\midrule

\multirow{8}{*}{\textbf{HM}}
& \multirow{2}{*}{DGI}
& Base     
& \besttab{11.67$\pm$0.06} & \besttab{3.11$\pm$0.06} & \besttab{15.78$\pm$0.32} & \besttab{14.20$\pm$0.23} & \besttab{5.53$\pm$0.10} & \besttab{67.22$\pm$1.10} \\
&
& +Tabular 
& 9.25$\pm$0.07 & 2.72$\pm$0.02 & 13.99$\pm$0.22 & 11.54$\pm$0.23 & 1.97$\pm$0.51 & 75.91$\pm$0.85 \\
\cmidrule(lr){2-9}

& \multirow{2}{*}{NS4GC}
& Base     
& \besttab{15.28$\pm$0.17} & \besttab{4.72$\pm$0.10} & \besttab{18.40$\pm$0.42} & \besttab{17.23$\pm$0.28} & 5.11$\pm$0.19 & 85.63$\pm$0.38 \\
&
& +Tabular 
& 15.13$\pm$0.33 & 4.24$\pm$0.12 & 18.10$\pm$0.47 & 17.17$\pm$0.54 & \besttab{7.43$\pm$1.07} & \besttab{70.28$\pm$1.49} \\
\cmidrule(lr){2-9}

& \multirow{2}{*}{DMoN}
& Base     
& \besttab{7.59$\pm$0.92} & \besttab{2.05$\pm$0.28} & 12.32$\pm$0.58 & \besttab{10.58$\pm$0.44} & \besttab{12.54$\pm$0.25} & \besttab{66.99$\pm$0.99} \\
&
& +Tabular 
& 6.28$\pm$1.06 & 1.60$\pm$0.68 & \besttab{14.41$\pm$0.85} & 9.76$\pm$1.27 & 5.77$\pm$6.36 & 69.12$\pm$7.65 \\
\cmidrule(lr){2-9}

& \multirow{2}{*}{Neuromap}
& Base     
& \besttab{7.44$\pm$0.17} & \besttab{1.45$\pm$0.07} & \besttab{12.73$\pm$0.25} & \besttab{9.37$\pm$0.18} & \besttab{6.29$\pm$0.17} & \besttab{20.28$\pm$0.45} \\
&
& +Tabular 
& 5.61$\pm$0.47 & 1.28$\pm$0.14 & 12.24$\pm$0.53 & 8.64$\pm$0.69 & 5.75$\pm$0.13 & 21.67$\pm$1.79 \\

\midrule

\multirow{8}{*}{\textbf{Pokec}}
& \multirow{2}{*}{DGI}
& Base     
& \besttab{4.20$\pm$0.01} & \besttab{0.32$\pm$0.00} & \besttab{2.23$\pm$0.02} & \besttab{1.59$\pm$0.01} & \besttab{6.65$\pm$0.04} & \besttab{92.40$\pm$0.03} \\
&
& +Tabular 
& 2.53$\pm$0.01 & 0.17$\pm$0.00 & 1.75$\pm$0.01 & 1.22$\pm$0.01 & 3.21$\pm$0.04 & 95.38$\pm$0.06 \\
\cmidrule(lr){2-9}

& \multirow{2}{*}{NS4GC}
& Base     
& \besttab{7.19$\pm$0.04} & \besttab{0.74$\pm$0.01} & \besttab{3.19$\pm$0.01} & \besttab{2.33$\pm$0.02} & \besttab{15.98$\pm$0.09} & \besttab{82.72$\pm$0.10} \\
&
& +Tabular 
& 3.57$\pm$0.01 & 0.29$\pm$0.00 & 2.06$\pm$0.03 & 1.40$\pm$0.02 & 7.81$\pm$0.04 & 90.69$\pm$0.04 \\
\cmidrule(lr){2-9}

& \multirow{2}{*}{DMoN}
& Base     
& \besttab{8.42$\pm$0.08} & \besttab{1.17$\pm$0.04} & \besttab{4.40$\pm$0.07} & \besttab{3.04$\pm$0.04} & \besttab{22.67$\pm$0.28} & 76.06$\pm$0.30 \\
&
& +Tabular 
& 6.55$\pm$0.56 & 0.89$\pm$0.11 & 4.16$\pm$0.32 & 2.20$\pm$0.26 & 21.81$\pm$2.45 & \besttab{74.30$\pm$5.83} \\
\cmidrule(lr){2-9}

& \multirow{2}{*}{Neuromap}
& Base     
& \besttab{8.53$\pm$0.25} & \besttab{1.84$\pm$0.11} & \besttab{5.12$\pm$0.21} & \besttab{2.62$\pm$0.14} & \besttab{31.71$\pm$0.56} & \besttab{65.39$\pm$0.65} \\
&
& +Tabular 
& 7.37$\pm$0.21 & 1.41$\pm$0.16 & 4.41$\pm$0.23 & 2.10$\pm$0.11 & 29.61$\pm$1.71 & 66.60$\pm$2.29 \\

\midrule

\multirow{6}{*}{\textbf{WebTopic}}
& \multirow{2}{*}{DGI}
& Base     
& 5.91$\pm$0.02 & 2.18$\pm$0.06 & 11.69$\pm$0.10 & \besttab{7.10$\pm$0.06} & -1.07$\pm$0.44 & \besttab{88.34$\pm$0.23} \\
&
& +Tabular 
& \besttab{7.14$\pm$0.14} & \besttab{2.55$\pm$0.06} & \besttab{13.50$\pm$0.29} & 5.84$\pm$0.08 & \besttab{1.40$\pm$0.27} & 91.75$\pm$0.28 \\
\cmidrule(lr){2-9}

& \multirow{2}{*}{NS4GC}
& Base     
& \besttab{10.05$\pm$0.04} & \besttab{3.86$\pm$0.05} & \besttab{15.69$\pm$0.14} & \besttab{8.80$\pm$0.15} & \besttab{24.51$\pm$0.06} & \besttab{52.27$\pm$0.69} \\
&
& +Tabular 
& 7.89$\pm$0.03 & 2.35$\pm$0.09 & 12.72$\pm$0.19 & 8.22$\pm$0.05 & 20.14$\pm$0.11 & 55.84$\pm$0.28 \\
\cmidrule(lr){2-9}

& DMoN
& +Tabular 
& \multicolumn{6}{c}{OOM} \\
\cmidrule(lr){2-9}

& Neuromap
& +Tabular 
& \multicolumn{6}{c}{OOM} \\

\bottomrule
\end{tabular}
\end{adjustbox}
\vspace{-0.2cm}
\end{table*}

\vpara{Observations.}
The results reveal that the benefit of neural tabular encoding is \emph{dataset- and method-dependent}, and does not uniformly improve over the \textsc{Base} protocol.

\begin{itemize}[leftmargin=*]
    \item \textbf{Limited gains on \texttt{HM} and \texttt{Pokec}.}
    On both datasets, \textsc{+Tabular} generally \emph{underperforms} the \textsc{Base} preprocessing across most methods and metrics. 
    For example, on \texttt{Pokec}, NS4GC's NMI drops from $7.19\%$ to $3.57\%$, and DGI's Modularity falls from $6.65\%$ to $3.21\%$.
    We attribute this to the relatively low number of features in these datasets (\texttt{HM}: 120 features; \texttt{Pokec}: 56 features), where quantile normalization already provides a sufficiently clean and stable input signal. 
    In this regime, introducing additional learnable parameters may increase optimization difficulty without a commensurate gain in representational power, particularly in the unsupervised setting where the clustering objective provides only a weak training signal.

    \item \textbf{Selective improvements on \texttt{WebTopic}.}
    In contrast, on \texttt{WebTopic} (528 features, $\mathcal{H}_e = 0.22$), \textsc{+Tabular} shows mixed but partially positive effects. 
    For DGI, the neural encoder improves NMI from $5.91\%$ to $7.14\%$ and ARI from $2.18\%$ to $2.55\%$, suggesting that learnable feature interactions are more beneficial when the raw feature space is higher-dimensional and more heterogeneous.
    However, NS4GC exhibits a consistent regression across all metrics, indicating that method-specific inductive biases interact non-trivially with the encoder design.

    \item \textbf{Scalability cost.}
    The neural tabular encoder introduces a non-trivial memory overhead. 
    For joint methods with dense pooling operations (DMoN, Neuromap), this overhead causes OOM errors on \texttt{WebTopic} (2.9M nodes, 528 features), highlighting a practical trade-off between model expressivity and scalability that is particularly acute for large industrial graphs.
\end{itemize}

\vpara{Recommendation.}
Based on these observations, we adopt the \textsc{Base} preprocessing protocol as the default in our main benchmark results, as it provides a more stable and universally applicable foundation across diverse dataset scales and model architectures.
The \textsc{+Tabular} encoder may offer targeted benefits for high-dimensional tabular graphs when paired with appropriate decoupled methods, but requires careful hyperparameter tuning and memory management.
We release both variants in \texttt{PyAGC} to allow practitioners to explore this trade-off in their specific deployment contexts.

\section{Miscellaneous}\label{app:misc}

\subsection{Limitations}
\label{app:limitation}
Although \texttt{PyAGC} aims to provide a comprehensive and scalable benchmark for attributed graph clustering, several limitations remain.
First, while our benchmark covers 12 datasets across different scales, domains, feature types, and homophily levels, it still cannot fully capture the diversity of industrial graph scenarios. Real-world deployments may involve dynamic graphs, temporal interactions, heterogeneous node and edge types, or private business-specific objectives that are not fully represented in the current benchmark.
Second, although we include structural metrics such as Modularity and Conductance, these metrics only capture certain topological properties of clusters. They may not always align with downstream business utility, semantic coherence, or human interpretability. Future work should incorporate task-specific and human-centered evaluation protocols.
Third, our scalability experiments are conducted on a single 32GB V100 GPU to ensure accessibility and fair comparison. While this setting demonstrates the practicality of mini-batch AGC methods, further engineering may be required for distributed training, streaming graphs, and billion-scale industrial deployment.

\subsection{Ethical Considerations}\label{app:ethical}
This work presents a benchmark and library for attributed graph clustering without collecting new human data or deploying models in real-world decision systems. All datasets used are publicly available or from prior benchmarks and are assumed to comply with existing licenses and anonymization practices.
Nevertheless, potential risks remain. First, included datasets (e.g., social or web graphs) may contain implicit biases or user-related signals, which could propagate to learned representations. Second, graph clustering techniques could be misused in applications such as surveillance or unfair profiling. We do not advocate such uses and encourage practitioners to follow relevant legal and ethical guidelines.
Finally, although \texttt{PyAGC} is designed for efficient training on a single GPU, large-scale experiments still incur non-trivial computational cost. We promote reproducibility through code and configuration release, while avoiding the distribution of sensitive data.

\subsection{Broader Impacts}\label{app:broader_impacts}
This work can have positive societal impacts by advancing scalable and robust graph clustering, which may benefit applications such as fraud detection, recommendation systems, and scientific discovery. In particular, improving clustering quality on large and complex graphs can help uncover meaningful structures in domains where labels are scarce.
However, these techniques also carry potential risks. When applied to user-centric data, they could enable large-scale profiling, surveillance, or unfair segmentation that may amplify existing biases or inequalities. Additionally, misinterpretation of clustering results could lead to harmful or unjustified decisions in sensitive applications. We therefore encourage practitioners to carefully consider fairness, privacy, and the broader societal implications when deploying these methods.

\end{document}